%% file: iclr2027_conference.tex
\documentclass{article} 
\usepackage{iclr2027_conference,times}
\iclrfinalcopy

\usepackage{amssymb}
\usepackage{algorithm}
\usepackage{algpseudocode}
\usepackage{xcolor}
\usepackage[most]{tcolorbox}
\usepackage{amsmath}
\usepackage{graphicx}
\usepackage{microtype}
\usepackage{hyperref}
\usepackage{pifont}
\usepackage{placeins}
\usepackage{url}
\usepackage{wrapfig}
\usepackage{listings}
\usepackage{enumitem}
\usepackage{booktabs}
\usepackage{multirow}
\usepackage{tabularx}
\usepackage{caption}
\usepackage{adjustbox}
\usepackage{xcolor}
\usepackage{colortbl}
\definecolor{gaingreen}{RGB}{0,128,0}
\definecolor{lossred}{RGB}{200,0,0}
\definecolor{oursrow}{RGB}{222,235,247}
\definecolor{blockgray}{gray}{0.92}
\newcommand{\gain}[1]{\raisebox{-0.35ex}{\tiny\textcolor{gaingreen}{+#1}}}

\definecolor{stratframe}{RGB}{70,105,150}
\newtcolorbox{promptbox}[1][]{
  enhanced,
  breakable,
  fontupper=\small\ttfamily,
  colback=gray!5,
  colframe=gray!50,
  boxrule=0.5pt,
  arc=3pt,
  left=8pt, right=8pt, top=8pt, bottom=8pt,
  #1
}
\definecolor{promptback}{RGB}{244,249,253}
\newtcolorbox{compactpromptbox}{
  enhanced, breakable,
  fontupper=\footnotesize\sffamily\raggedright,
  colback=promptback, colframe=stratframe!70,
  boxrule=0.5pt, arc=2pt,
  left=6pt, right=6pt, top=6pt, bottom=6pt,
  before skip=4pt, after skip=6pt
}

\usepackage{xspace}
\usepackage[T1]{fontenc}

\definecolor{darkblue}{rgb}{0, 0, 0.5}
\hypersetup{colorlinks=true, citecolor=darkblue, linkcolor=darkblue, urlcolor=darkblue}

\newcommand{\policyname}{self-improvement policy\xspace}
\newcommand{\policynames}{self-improvement policies\xspace}
\newcommand{\algcomment}[1]{\hfill{\footnotesize$\triangleright$~#1}}
\newcolumntype{Y}{>{\centering\arraybackslash}X}

\title{\textsc{Meta-TTL}: Meta-Learning Self-Improvement \\ Policies for Language Agents}

\author{Zhanzhi Lou\textsuperscript{1}, Hui Chen\textsuperscript{1}, Yibo Li\textsuperscript{1}, Qian Wang\textsuperscript{1}, Bryan Hooi\textsuperscript{1} \\
\textsuperscript{1}National University of Singapore \\
\texttt{\{hui.chen,dcsbhk\}@nus.edu.sg}}

\newcommand{\modelname}{\textsc{Meta-TTL}\xspace}

\begin{document}

\maketitle
\lhead{Preprint}

\begin{abstract}
    Test-Time Learning (TTL) enables language agents to iteratively refine their performance through repeated interactions with the environment at inference time. At the core of TTL is a \policyname that updates the actor policy based on experience from previous episodes, thereby improving future behavior. Existing methods rely on hand-crafted \policynames rather than optimizing them for downstream improvement. We argue that optimal \policynames should be learned from task environments, not hand-engineered based on human intuition. To achieve this, we introduce \textbf{\modelname}, a framework that formulates the discovery of effective \policynames as a bi-level optimization problem. Within this framework, the inner loop executes the standard TTL process, measuring how effectively a candidate \policyname helps an agent correct errors across sequential episodes. Guided by the agent's performance, the outer loop performs reflective meta-training across diverse
    training tasks, using a balanced improvement score (BIS) to balance task contributions during candidate selection. We evaluate \modelname on Jericho, WebArena-Lite, and $\tau^2$-bench across both in-distribution (ID) and out-of-distribution (OOD) settings.  \modelname consistently outperforms existing baselines, improving TTL over the strongest baseline by up to 23\% on ID tasks and 27\% on OOD tasks. These results suggest that the optimized \policyname encodes transferable meta-strategies that generalize beyond the training task distribution. Code is available at \url{https://github.com/zzzlou/meta-ttl}.

\end{abstract}

\section{Introduction}
\begin{figure}[h]
    \centering
    \includegraphics[width=\textwidth]{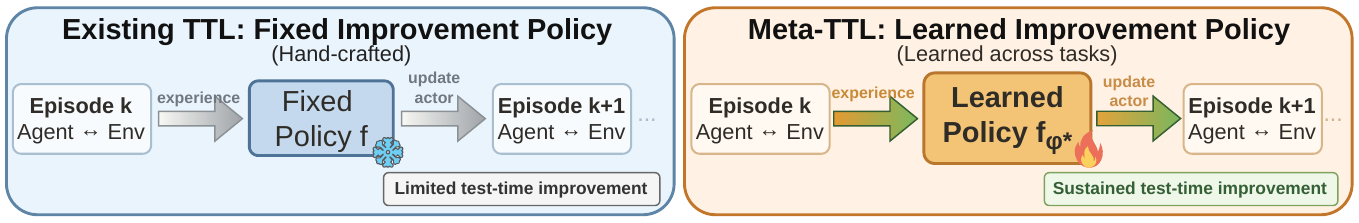}
    \caption{Self-improvement policies determine how the agent uses experience up to episode $k$ to update the actor before episode $k+1$. Existing methods use a hand-crafted \policyname, whereas \modelname learns it offline across tasks and applies it at test time.}
    \label{fig:intro}
\end{figure}
Large Language Model (LLM) agents have demonstrated strong zero-shot capabilities across a wide range of tasks. In practice, however, agents deployed in novel environments often struggle to adapt on the fly \citep{gao2026surveyselfevolvingagentswhat,fang2025comprehensivesurveyselfevolvingai}. Consider a human player encountering an unfamiliar video game: they fail, diagnose what went wrong, adjust their strategy, and try again, often improving with each iteration. This capacity for Test-Time Learning (TTL), the ability to accumulate experience over repeated interactions and achieve progressively better performance \citep{wu2024streambenchbenchmarkingcontinuousimprovement,he2025evotestevolutionarytesttimelearning,wei2025evomemorybenchmarkingllmagent}, remains limited in current LLM agents. Without parameter updates or ground-truth supervision, they often treat every episode as an independent zero-shot trial, repeating the same errors regardless of how many attempts they are given \citep{jiang2026adaptationagenticaisurvey}.

At the core of TTL is a \emph{\policyname} that updates the actor policy based on accumulated experience. Unlike the actor policy, which determines the agent's behavior within an episode, the \policyname determines how the actor policy evolves across episodes. However, most existing methods, such as Reflexion \citep{shinn2023reflexionlanguageagentsverbal}, perform self-improvement by relying purely on the pretrained capabilities of the underlying LLM. Fundamentally, the \policyname serves as a learning algorithm: it maps past experience to future behavioral improvement. Such capabilities require dedicated optimization \citep{thrun1998learning,10.5555/216000.216006} that general-purpose language modeling does not provide \citep{radford2019language,li2024hindsight2020testinglimits,brown2020languagemodelsfewshotlearners}.

In this work, we take the view that \textbf{effective test-time self-improvement is itself a learnable capability rather than a byproduct of a general-purpose LLM} \citep{liu2025position}. Instead of hand-engineering the agent's cross-episode \policyname, we seek to \textit{learn} it from task environments by optimizing it for downstream improvement at test time.

To this end, we propose \textbf{\modelname}, a framework that treats learning to perform TTL effectively as a meta-learning problem: given a distribution of training tasks, we formulate the discovery of effective \policynames as a bi-level optimization. Concretely, this bi-level structure consists of an inner TTL loop and an outer meta-training loop. In the inner loop, an LLM agent interacts with the environment over a series of episodes and adapts based on prior attempts, measuring how well a candidate \policyname $\phi$ helps the agent improve across episodes. In the outer loop, we optimize $\phi$ over a distribution of training tasks through reflective meta-training: we iteratively refine candidate policies, evaluate them through the inner loop, and retain those that produce stronger TTL performance. At test time, the learned \policyname is frozen and applied to unseen tasks.

A key distinction from prior work lies in what is being optimized (Figure~\ref{fig:intro}). Existing TTL methods treat the \policyname (how the actor policy is updated between episodes) as a hand-designed component, and focus on improving the actor's behavior within a single task session through ad-hoc verbal feedback \citep{shinn2023reflexionlanguageagentsverbal,madaan2023selfrefineiterativerefinementselffeedback} or memory accumulation \citep{packer2024memgptllmsoperatingsystems,xu2025amemagenticmemoryllm}. We instead treat the \policyname itself as the object of optimization: \modelname learns how an agent should improve across episodes from a distribution of training tasks, and deploys the resulting \policyname at test time.

Our contributions are as follows: \begin{itemize} 

\item We formulate learning an effective \policyname for TTL as a \textbf{meta-learning problem}, providing a principled framework for optimizing how agents improve themselves across episodes. 
\item We propose \textbf{\modelname}, which learns a \policyname across a distribution of training tasks through reflective meta-training and deploys it on unseen tasks. The learned policy is instantiated as a natural-language meta-prompt. 
\item We evaluate \modelname on three language-based sequential decision-making benchmarks. \modelname achieves the highest average W-AUC on every benchmark in both ID and OOD settings, improving over the strongest baseline by 10--23\% on ID tasks and 5--27\% on OOD tasks. \end{itemize}

\section{Related Work}

\paragraph{Test-Time Learning} \label{sec:related_ttl_def}
Test-Time Learning (TTL) improves agents after deployment by learning from their own experience \citep{jiang2026adaptationagenticaisurvey}. \textbf{Gradient-based} methods update model weights at test time via fine-tuning \citep{akyurek2025the,acikgoz2025selfimprovingllmagentstesttime,zweiger2025selfadapting,ye2026onlineexperientiallearninglanguage} or reinforcement learning \citep{zuo2025ttrl,yuksekgonul2026learningdiscovertesttime}. \textbf{Weight-frozen} methods instead adapt through external state. Some accumulate experience through verbal reflection \citep{shinn2023reflexionlanguageagentsverbal,madaan2023selfrefineiterativerefinementselffeedback}, memory \citep{wang2024agentworkflowmemory,wei2025evomemorybenchmarkingllmagent,chhikara2025mem0buildingproductionreadyai,suzgun2025dynamiccheatsheettesttimelearning,zhou2025mementofinetuningllmagents,xu2025amemagenticmemoryllm}, or insights and rules distilled from experience \citep{zhao2024expelllmagentsexperiential,majumder2023clincontinuallylearninglanguage,chen2026grounded,zhang2025agentlearningearlyexperience}. Others search over prompts using LLM proposals \citep{zhou2023largelanguagemodelshumanlevel,yang2024largelanguagemodelsoptimizers}, evolutionary search \citep{fernando2023promptbreederselfreferentialselfimprovementprompt,guo2025evopromptconnectingllmsevolutionary,ye2024reevolargelanguagemodels}, or natural-language feedback \citep{yuksekgonul2024textgradautomaticdifferentiationtext,gupta2024metareflectionlearninginstructionslanguage,zhang2026agenticcontextengineeringevolving}. This line of work has been further extended to compound AI systems and scientific discovery \citep{agrawal2026gepa,novikov2025alphaevolvecodingagentscientific,liu2026evoxmetaevolutionautomateddiscovery} and to evolving the entire agent configuration \citep{he2025evotestevolutionarytesttimelearning}. In all of these methods, however, the goal is to \emph{improve task-specific behavior} under a hand-crafted \policyname, whereas \modelname aims to \emph{learn the \policyname itself} from a distribution of training tasks.

\paragraph{Meta-Learning}
Meta-learning seeks to extract transferable knowledge from a task 
distribution so that a learner can adapt efficiently to new tasks 
\citep{thrun1998learning,hospedales2021meta}. In the context of LLMs, 
in-context learning \citep{dong2024surveyincontextlearning} has been viewed as a form of black-box meta-learning, where 
adaptation arises through context conditioning rather than weight 
updates \citep{brown2020languagemodelsfewshotlearners,dherin2025learningtrainingimplicitdynamics}. Earlier work such as STaR \citep{zelikman2022star} and SCoRe \citep{kumartraining} explicitly optimizes self-improvement through self-generated rationales or RL-based self-correction, but does not learn cross-episode \policynames for sequential environments. Several 
concurrent works train self-improvement capabilities via RL: LAMER~\citep{jiang2026metarlinducesexplorationlanguage} meta-trains exploration strategies, MR-Search~\citep{xiao2026metareinforcementlearningselfreflectionagentic} 
learns cross-episode self-reflection, and 
LSE~\citep{chen2026learningselfevolve} trains a prompt-editing policy 
with a single-step objective. All three require fine-tuning model 
weights via policy gradients. In contrast, our framework operates entirely in 
prompt space through gradient-free search, yielding an interpretable text 
artifact.

\section{Methodology}
\label{sec:method}

\begin{figure}[!t]
    \centering
    \includegraphics[width=\textwidth]{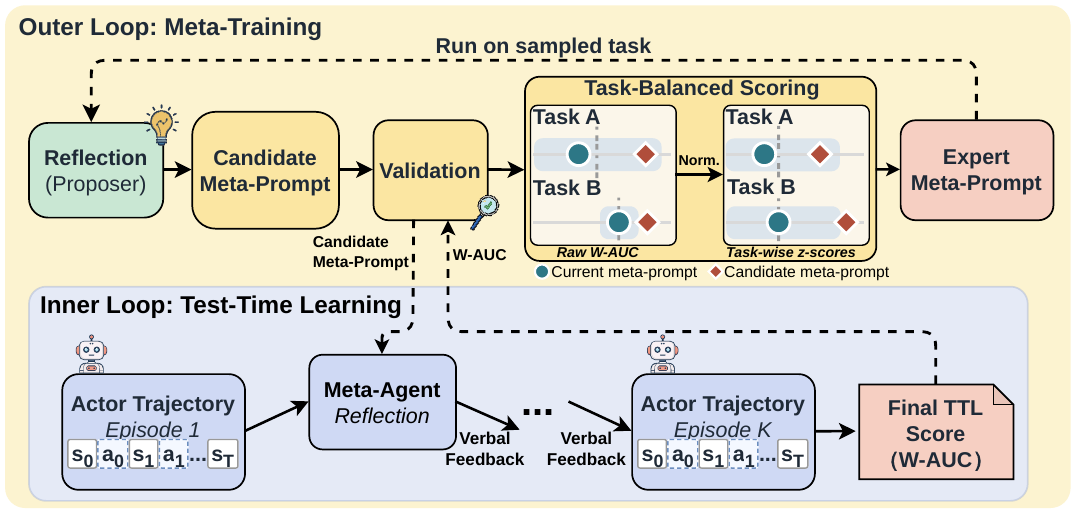}
    \caption{
Overview of \modelname.
\textbf{Outer loop (meta-training):}
A proposer reflects on a TTL session generated by the current
expert meta-prompt to propose a candidate meta-prompt.
The candidate is evaluated across validation tasks and replaces
the expert if its balanced improvement score (BIS) is positive.
\textbf{Inner loop (test-time learning):}
Given a meta-prompt, the meta-agent generates verbal feedback
from the actor's trajectories to rewrite its system prompt
between episodes. The resulting TTL session is scored using W-AUC.
}

    \label{fig:illustration}
\end{figure}

We present \textbf{\modelname}, a bi-level framework for learning a \policyname for test-time learning in language agents. As shown in Figure~\ref{fig:illustration}, \modelname couples an inner TTL loop with an outer meta-training loop. The inner loop adapts the actor across episodes by rewriting its system prompt, while the outer loop improves the meta-prompt by proposing candidates from rollouts and retaining the candidate only if it improves on validation tasks.

\subsection{Test-Time Learning Formulation}
\label{sec:ttl_formulation}

We model each task instance $g$ as a finite-horizon Partially Observable Markov Decision Process (POMDP),
\[\mathcal{M}_g = (\mathcal{S}, \mathcal{A}, \mathcal{T}, \Omega, \mathcal{R}, H),
\]
where $\mathcal{S}$ is the latent state space, $\mathcal{A}$ is the action space, $\mathcal{T}$ is the transition kernel, $\Omega$ is the observation space, $\mathcal{R}$ is the task-specific reward function, and $H$ is the episode horizon.
Here, $g$ denotes a single task instance, such as one Jericho game or one WebArena task.

A \textbf{TTL session} on task $g$ consists of $K$ consecutive episodes, denoted by $\xi_g = (\tau_1, \tau_2, \dots, \tau_K)$. After each episode, the environment resets to its initial state, so improvement across the session must come from adaptation in the agent rather than from environmental state continuity. We score a session using Weighted Area Under the Learning Curve (W-AUC): To distinguish improving from declining performance that uniform averaging would treat equally, we assign greater
weight to later episodes. We adopt linear weights, $w_k=k$, motivated by the approximately linear accumulation of experience available to the meta-agent:
\begin{equation}
\label{eq:wauc}
\text{W-AUC}(\xi_g) = \frac{\sum_{k=1}^{K} w_k \cdot J(\tau_k)}{\sum_{k=1}^{K} w_k \cdot J_{\max}(g)}, \quad w_k = k
\end{equation}
where $\tau_k$ is the trajectory of episode $k$, $J(\tau_k)$ is its reward, and $J_{\max}(g)$ is the maximum achievable return for task $g$.

\subsection{Learnable Self-Improvement Policies}
\label{sec:framework}

In a TTL session, two distinct policies interact. The actor policy $\pi$ determines behavior within a single episode, selecting actions given the current observation. The \policyname $f$ operates at a higher level: after each episode, it observes the accumulated experience and produces an updated actor policy for the next attempt. That is,
\begin{equation}
\label{eq:adaptation_general}
\pi_{k+1} = f(\pi_k, \mathcal{H}_k),
\end{equation}
where $\mathcal{H}_k = \{\tau_1, \dots, \tau_k\}$ is the trajectory history up to episode $k$. Existing TTL methods typically hand-design $f$ (e.g., a fixed reflection prompt, carefully designed memory system). Our goal is to learn $f$ from a distribution of training tasks.

In general, an LLM-based actor policy is jointly determined by its weights $\theta$ and its prompt $c$. The \policyname can therefore operate along two axes: modifying $\theta$ (gradient-based adaptation) or modifying $c$ (prompt-based adaptation). We focus on the prompt-based instantiation, where $\theta$ is frozen and all behavioral change is mediated through system prompt rewriting. This avoids gradient computation at test time and makes adaptation lightweight.

\textbf{Actor.}
A frozen LLM $\pi_{\theta}$ interacts with the environment. We designate its system prompt $\rho$ as the modifiable component of the context: in episode $k$, the actor executes $\tau_k \sim \pi_{\theta}(\cdot \mid \rho_k)$. Since $\theta$ is fixed, updating $\rho$ is the sole mechanism for changing the actor's behavior across episodes.

\textbf{Meta-Agent.}
We instantiate the \policyname $f$ as a separate LLM governed by a meta-prompt $\phi$. After episode $k$, the meta-agent observes the trajectory history $\mathcal{H}_k$ and generates the updated system prompt:
\begin{equation}
\label{eq:adaptation_prompt}
\rho_{k+1} \sim f_{\phi}(\cdot \mid \rho_k, \mathcal{H}_k)
\end{equation}
The meta-prompt $\phi$ serves as the \policyname: it determines what aspects of past experience the meta-agent attends to, how it diagnoses failures, and what form of guidance it produces. The learnable component is therefore $\phi$. Rather than hand-crafting it or relying on fixed heuristics, we optimize it through meta-training (\S\ref{sec:training}).

\subsection{Reflective Meta-Training}
\label{sec:training}

The goal of meta-training is to find a meta-prompt $\phi^*$ that maximizes expected TTL performance on the training tasks:
\begin{align}
\phi^* = \operatorname*{argmax}_{\phi} \mathbb{E}_{g \sim \mathcal{D}_{\text{train}}} \left[ \text{W-AUC}(\xi_g^{\phi}) \right].
\end{align}
where $\xi_g^\phi$ denotes the TTL session on task $g$ run with meta-prompt $\phi$.

\paragraph{Overview} As shown in the outer loop of Figure~\ref{fig:illustration}, meta-training maintains a single expert meta-prompt $\phi^\star$, initialized from the seed meta-prompt $\phi_0$. At each iteration, we sample a task from the train set and run it using $\phi^\star$ to produce a full TTL session. A reflection module then reflects on the resulting TTL session to propose a revised meta-prompt candidate $\tilde{\phi}$, which is compared against the current expert meta-prompt over the validation set. The candidate replaces $\phi^\star$ if its balanced improvement score (defined below) is positive.

\textbf{Two-Stage Reflection.}
Given a TTL session $\xi_g$ generated under the current meta-prompt $\phi^\star$, the first stage extracts task-specific facts $F_g$ from the session trajectory. In the second stage, the reflection model uses these
facts alongside $\xi_g$ and $\phi^\star$ to identify issues in the meta-agent's self-improvement strategy and produce a revised meta-prompt $\tilde{\phi}$.

\textbf{Validation.}
Each proposed meta-prompt is evaluated on a set of validation tasks by running TTL sessions using a meta agent, whose guiding-behavior is governed by the meta-prompt. For every validation task $h \in \mathcal{D}_{\mathrm{val}}$, the corresponding validation score of a meta-prompt candidate $\phi$ is:
\begin{align}
s_{h}^\phi &= \mathrm{W\text{-}AUC}(\xi_{h}^\phi)
\end{align}

\textbf{Task-Balanced Scoring and Selection.}
Although W-AUC normalizes each task’s returns by its maximum achievable reward, some tasks can still be intrinsically easier to improve than others. A change in meta-prompt may thus produce a much larger W-AUC difference on these tasks, giving them disproportionate influence if scores are directly averaged across tasks. To mitigate this imbalance, we define a balanced
improvement score (BIS) by normalizing each task's W-AUC against
its score history $\mathcal{H}_{h,t}$ and averaging the candidate's normalized
gains over the current expert across tasks:
\begin{align}
z_{h,t}^{\phi} &=
\frac{s_h^\phi-\mu(\mathcal{H}_{h,t})}
{\sigma(\mathcal{H}_{h,t})},\\
\mathrm{BIS}_t(\tilde{\phi},\phi^\star) &=
\frac{1}{|\mathcal{D}_{\mathrm{val}}|}
\sum_{h\in\mathcal{D}_{\mathrm{val}}}
\left(z_{h,t}^{\tilde{\phi}}-z_{h,t}^{\phi^\star}\right).
\end{align}
Both meta-prompts are normalized using the same history of scores.
The candidate $\tilde{\phi}$ replaces the
current expert $\phi^\star$ only if
$\mathrm{BIS}_t(\tilde{\phi},\phi^\star)>0$. More details can be found in Appendix \ref{app:algo_descriptions}

\subsection{Evaluation}
\label{sec:inference}

After offline meta-training, we freeze the learned meta-prompt
$\phi^\star$ and evaluate it on in-distribution (ID) and out-of-distribution (OOD) tasks. ID and OOD are defined relative to meta-training: ID tasks come from domains used
during meta-training, whereas OOD tasks come from held-out domains. For WebArena-Lite and $\tau^2$-bench, ID evaluation uses
held-out task instances within the seen domains. For Jericho, where each game is a single task, ID evaluation uses the meta-training games. Within each evaluation task, the meta-agent uses the fixed
$\phi^\star$ to update the actor's prompt between episodes.

\section{Experiments}
\label{sec:exp}

\subsection{Experimental Setup}
\label{sec:exp_setup}

\textbf{Benchmarks.}
We evaluate \modelname on three benchmarks: \textbf{Jericho} \citep{hausknecht2020interactivefictiongamescolossal}, a suite of interactive fiction games; \textbf{WebArena-Lite} \citep{zhou2024webarenarealisticwebenvironment}, a web-navigation benchmark; and \textbf{$\tau^2$-bench} \citep{barres2025tau2benchevaluatingconversationalagents}, a realistic customer-service benchmark involving conversational tool use. For Jericho, Detective, Zork~1, and Temple are used for meta-training
and ID evaluation, while Balances, Library, and Zork~3 are held out for OOD evaluation. For WebArena-Lite, Shopping, GitLab, and Map are ID domains, whose tasks are split into training, validation, and evaluation sets; Reddit and Shopping Admin are OOD domains. For $\tau^2$-bench, Airline and Retail are ID domains, and Telecom is OOD. Each TTL session contains six episodes on Jericho and three on WebArena-Lite and $\tau^2$-bench.

\textbf{Models and Baselines.}
All main experiments use a frozen \textbf{Gemini-3-Flash} actor. We compare against Static as a non-learning baseline;
Reflexion \citep{shinn2023reflexionlanguageagentsverbal}, a Memory Agent \citep{he2025evotestevolutionarytesttimelearning} that retains cross-episode histories, and AWM \citep{wang2024agentworkflowmemory} as reflection- and memory-based methods; and TextGrad \citep{yuksekgonul2024textgradautomaticdifferentiationtext}, EvoPrompt \citep{guo2025evopromptconnectingllmsevolutionary}, and EvoTest \citep{he2025evotestevolutionarytesttimelearning} as prompt- and
configuration-optimization methods. Methods that use a separate LLM to update the actor between episodes (the \emph{adaptation LLM}, corresponding to the meta-agent in \modelname) use GPT-5, the same model as \modelname's meta-agent. We also include a \textbf{Naive} meta-agent that shares the same architecture and models of \modelname but uses an unoptimized \policyname. We report two metrics for each TTL session: \textbf{Average Score}, the mean score across episodes, and \textbf{W-AUC} (Eq.~\ref{eq:wauc}). Results with additional model backbones are reported in Appendix~\ref{app:backbones}.

\input{tables/jericho}
\input{tables/webarena}

\subsection{Main Results}
\label{sec:results}

\begin{wrapfigure}{R}{0.5\textwidth}
    \centering
    \vspace{-14pt}
    \includegraphics[width=0.5\textwidth]{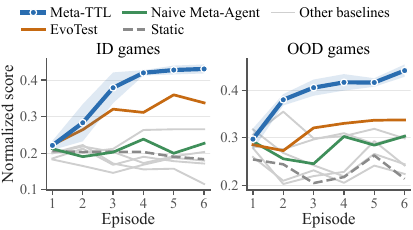}
    \vspace{-16pt}
    \caption{Per-episode normalized score on Jericho, averaged over ID and OOD games. \modelname shows the largest improvement over the session. Shading denotes the standard error over five runs.}
    \label{fig:learning_curves}
    \vspace{-8pt}
\end{wrapfigure}

\textbf{\modelname consistently improves test-time learning across all three benchmarks.} Under matched model backbones, \modelname achieves the highest average W-AUC on the ID tasks of all three benchmarks  (Tables~\ref{tab:jericho_id}, \ref{tab:webarena_id}, and~\ref{tab:tau2bench}),  improving over the strongest baseline by 23\% on Jericho, 12\% on WebArena-Lite, and over 10\% on both the Airline and Retail domains of $\tau^2$-bench. These gains show that the benefit of \modelname extends beyond text games to web navigation and conversational tool-use. \textbf{The learned \policyname also generalizes to out-of-distribution tasks}, achieving the highest W-AUC on OOD games and domains, including unseen websites in WebArena-Lite and the Telecom domain of $\tau^2$-bench (Tables~\ref{tab:jericho_ood}, \ref{tab:webarena_ood}, and~\ref{tab:tau2bench}). As shown in Figure~\ref{fig:learning_curves}, \modelname's advantage on Jericho grows over the session, with the largest improvement across episodes on both ID and OOD games. Together, these results indicate that \modelname's meta-learned self-improvement policy achieves better test-time learning performance than hand-designed heuristic methods, and that it transfers beyond the training tasks. We analyze what the learned policies encode and why they help in Section~\ref{sec:rq3}.
\input{tables/tau2bench}

\input{tables/parameter_ttl}

\paragraph{\modelname outperforms parameter-based TTL methods.}
We compare \modelname with Online SFT, Online GRPO \citep{shao2024deepseekmathpushinglimitsmathematical, he2025evotestevolutionarytesttimelearning}, and SFT+MAML \citep{finn2017modelagnosticmetalearningfastadaptation,nichol2018firstordermetalearningalgorithms} on Jericho. All methods use a Qwen3-8B LLM backbone (see Appendix~\ref{app:parameter_ttl} for details). As shown in Table~\ref{tab:parameter_ttl}, \modelname achieves the highest average W-AUC in both ID and OOD settings and improves consistently over the static actor on all six games. The parameter-updating methods yield smaller average gains and are unstable across games, suggesting that they struggle to learn reliably from only a few sparse-reward episodes.

\subsection{Ablation Study}
\label{sec:rq2}

\paragraph{Reflective meta-training matches RL-based training with far fewer rollouts.}
We provide a comparison between training the same Qwen3-4B meta-agent by GRPO \citep{shao2024deepseekmathpushinglimitsmathematical} and by \modelname's outer-loop, both using a shared frozen Gemini-3-Flash actor. GRPO updates the meta-agent's weights, whereas our reflective meta-training updates its meta-prompt. As shown in Table~\ref{tab:outer_loop_ablation}, the \modelname outer loop matches or exceeds GRPO on both ID and OOD averages while using far fewer rollouts (350 vs.\ 1{,}920). Moreover, \modelname's outer loop is gradient-free and produces a readable meta-prompt rather than model weights.

\paragraph{\modelname's outer loop outperforms GEPA.}
We also compare with GEPA \citep{agrawal2026gepa}, a widely used reflective prompt optimizer that maintains a Pareto frontier of candidates across validation tasks. As shown in Table~\ref{tab:outer_loop_ablation}, GEPA performs well on some ID games but transfers less effectively to OOD games. A plausible explanation is that GEPA's candidate-selection mechanism preserves prompts specialized to individual tasks, which is not well aligned with the meta-learning objective of finding a single \policyname that generalizes. This is also consistent with our finding that the prompts produced by GEPA contain substantial task-specific facts and solutions, which can benefit seen tasks but are less useful on unseen ones. Full meta-prompts can be found in Appendix~\ref{app:optimized_prompts}.

\input{tables/outer_loop_ablation}

\subsection{Analysis}
\label{sec:rq3}

\input{tables/strategies}

\paragraph{Emergent self-improvement strategies.}
Examining the meta-prompts learned by \modelname (Figure~\ref{fig:strategies}), we find that they share several recurring strategies. These include \textbf{(i) meta-level credit assignment}: attributing a failure either to the actor not following the guidance or to the guidance itself being wrong; \textbf{(ii) verified knowledge retention}: retaining verified findings together with the conditions under which they hold; \textbf{(iii) bounded exploration}: preserving the best verified behavior while exploring only through small changes. Collectively, these strategies describe \textbf{how to learn from previous attempts rather than how to solve any particular task}, suggesting that they operate at the meta level and target the TTL process itself. Appendix~\ref{app:case_studies} provides case studies of how these strategies lead to improvement at test time.

\paragraph{Efficiency analysis.}
\emph{Test time.} As shown in Table~\ref{tab:efficiency}, \modelname's test-time latency is comparable to existing prompt-optimization baselines. It uses five adaptation LLM calls per six-episode Jericho session, with total LLM calls and token usage within the range of these baselines.
\emph{Meta-training.} Meta-training incurs a one-time offline cost that requires no GPU training. Across the three benchmarks, meta-training runs for 12--15 iterations with 205--350 TTL sessions and about \$30--70 in API cost per benchmark, depending on the benchmark and model backbone. The learned meta-prompt is then applied to all test tasks without further training.
\input{tables/efficiency}

\FloatBarrier

\section{Conclusion}
\label{sec:conclusion}
This paper introduces \textbf{\modelname}, a bi-level framework that learns \policynames for test-time learning in language agents through reflective meta-training across tasks. We find that the learned \policyname consistently outperforms hand-crafted and unoptimized baselines on Jericho, WebArena-Lite, and $\tau^2$-bench, with gains extending to out-of-distribution tasks. These results suggest that how agents self-improve from experience is itself a learnable ability, and that test-time learning may benefit from meta-learning the self-improvement procedure. 




\bibliography{iclr2027_conference}
\bibliographystyle{iclr2027_conference}

\appendix

\section{Optimized Meta-Prompts}
\label{app:optimized_prompts}

The complete optimized meta-prompts for Jericho, WebArena-Lite, and $\tau^2$-bench used in the main experiments are shown below.

\subsection{Jericho Meta-Prompt}
\label{app:jericho_prompt}
\begin{compactpromptbox}
You are the META agent in a test-time-learning loop for parser-based text adventures. Read the complete task history and write replacement instructions that maximize the ACTOR's expected game score in the next fresh episode.\par
\par\vspace{0.2em}
Optimize the session's time-weighted normalized score AUC. Favor early reliable points, survival or victory, preservation of strong runs, and low-regret improvement. Exploration is valuable only when its expected future benefit justifies its immediate turn cost and risk.\par
\par\vspace{0.2em}
Every episode starts from the initial state. Guidance must be self-contained: carry forward learned commands, routes, prerequisites, hazards, parser behavior, and uncertainty, but never imply that inventory, location, opened passages, NPC state, defeated threats, or other world state persists.\par
\par\vspace{0.2em}
\textbf{Diagnose adaptation from executed evidence:}
\begin{itemize}[leftmargin=1.25em,itemsep=0pt,topsep=1pt,parsep=0pt,partopsep=0pt]
\item Compare the latest episode with the best prior episode by final score, timing of points, survival or victory, useful discoveries, and turns wasted.
\item Reconstruct commands actually executed and their resulting observations. Do not treat requested or intended actions as executed.
\item Identify which prior recommendations were followed, ignored, attempted in the wrong state, blocked by parser syntax, or made impossible by missing prerequisites.
\item Credit useful actor improvisations and promote them only under the source conditions where they succeeded.
\item Ignored advice remains untested; do not claim it succeeded or failed.
\item After regression or stagnation, identify the smallest new action, omission, ambiguity, mistimed trigger, or speculative branch responsible.
\item If compliance was poor, simplify the plan and place the critical command directly beside its observation trigger rather than repeating more explanation.
\end{itemize}
\par\vspace{0.2em}
\textbf{Maintain an evidence-based operational model:}
\begin{itemize}[leftmargin=1.25em,itemsep=0pt,topsep=1pt,parsep=0pt,partopsep=0pt]
\item Record verified scoring actions, automatic rewards, milestones, survival and terminal outcomes.
\item Preserve the contiguous executed trajectory leading to each milestone.
\item Track location, inventory, NPC and event state, prerequisites, one-way changes, timers, hazards, and return routes.
\item Record exact successful, failed, and partially parsed commands.
\item Separate direct observations from hypotheses.
\item Estimate turn costs from executed commands, including travel, acquisition, retries, and scripted delays.
\end{itemize}
\par\vspace{0.2em}
\textbf{Treat state precisely:}
\begin{itemize}[leftmargin=1.25em,itemsep=0pt,topsep=1pt,parsep=0pt,partopsep=0pt]
\item Model a transition as source observation and full relevant state, entered command, resulting observation, and resulting state.
\item Relevant state includes scripted-event progress, NPC presence and readiness, object accessibility, inventory, open or locked passages, and any recent state-changing event.
\item A command is verified only from the source state in which it succeeded.
\item A room name or visible object is not by itself proof that all required state conditions hold.
\item If the same command fails and later succeeds, compare the two source states. Trigger it only on the weakest distinguishing cue directly supported by the log.
\item Never generalize an NPC-dependent command across earlier dialogue, wake-up, following, hostility, or cooperation states without evidence.
\item Account for automatic movement and cutscenes. After one occurs, infer the current location only from the resulting observation, not from the command that initiated it.
\item Repeated room names are not interchangeable; distinguish them by text, exits, objects, NPCs, and events.
\end{itemize}
\par\vspace{0.2em}
\textbf{Reconstruct routes conservatively:}
\begin{itemize}[leftmargin=1.25em,itemsep=0pt,topsep=1pt,parsep=0pt,partopsep=0pt]
\item An advertised exit proves only that it is described, not its destination, safety, reversibility, reward, or prerequisites.
\item Never assume an opposite direction returns.
\item Never splice transitions observed in incompatible states.
\item Base the primary route on an actor trajectory that actually succeeded.
\item Before shortening or modifying a route, replay every retained transition against the log.
\item Remove a detour only when the log verifies both the resulting state and the next command from that state.
\item Do not predict saved turns, safety, points, or success unless the modified trajectory supports the claim.
\end{itemize}
\par\vspace{0.2em}
\textbf{Protect reliable baselines:}
\begin{itemize}[leftmargin=1.25em,itemsep=0pt,topsep=1pt,parsep=0pt,partopsep=0pt]
\item Treat the best surviving or winning trajectory as the default.
\item Preserve its scoring prefix and verified finish unless evidence supports a specific local edit.
\item Prefer one verified local improvement over a broad rewrite.
\item Reproduce reliable points as early as practical using only state-valid changes.
\item Preserve profitable detours when observed reward justifies their executed cost.
\item Do not replace a strong baseline with speculative shortcuts or several unrelated branches.
\item Once a strong route exists, secure its valuable prefix before optional tests unless a test must occur earlier.
\end{itemize}
\par\vspace{0.2em}
\textbf{Make guidance executable:}
\begin{itemize}[leftmargin=1.25em,itemsep=0pt,topsep=1pt,parsep=0pt,partopsep=0pt]
\item Tie every important command to a distinctive observation or verified state cue.
\item At ambiguous or repeated locations, specify the cue that distinguishes the correct instance.
\item Do not use "continue," "keep going," "go until," "follow the route," "explore," or "exhaust interactions."
\item Do not use blanket directional rules when intermediate choices or state changes exist.
\item Do not tell the ACTOR to "stop" unless a verified enterable command intentionally ends the game. Assume the ACTOR may continue until the step limit.
\item After every focused test, specify what to do for each important observed result and what productive action follows afterward.
\item Ensure the plan covers the remaining useful episode budget rather than ending at an early discovery.
\item Use one internally consistent primary route. Do not add summaries that abbreviate or contradict it.
\item Keep hazard warnings and corrections beside the junction where they apply.
\item Avoid large menus of guessed commands.
\end{itemize}
\par\vspace{0.2em}
\textbf{Handle failures and recovery:}
\begin{itemize}[leftmargin=1.25em,itemsep=0pt,topsep=1pt,parsep=0pt,partopsep=0pt]
\item Do not respond to a failed state-dependent command with syntax variants until prerequisites and source state have been checked.
\item If a command succeeds elsewhere, route back to its verified source state instead of repeatedly retrying it.
\item Permit a retry only when an observed state change could materially alter the outcome.
\item After repeated parser failures, missing prerequisites, or a loop, abandon that line and execute a specified alternative.
\item Detect loops by repeated state-action patterns without new score, state change, or information. Explicitly prohibit repeating the loop and choose a different verified transition.
\item Never invent a recovery or retreat route.
\end{itemize}
\par\vspace{0.2em}
\textbf{Choose exploration deliberately:}
\begin{itemize}[leftmargin=1.25em,itemsep=0pt,topsep=1pt,parsep=0pt,partopsep=0pt]
\item Treat unvisited exits and untested interactions as potentially lethal, trapping, timed, one-way, or costly.
\item Do not infer safety or reward from names, narrative tone, genre conventions, visible objects, or apparent puzzle relevance.
\item Describe untested actions as tests, not as likely rewards or likely solutions.
\item Prefer one focused frontier near the latest safely reached productive state.
\item A focused test must have a clear question, a bounded number of commands, observable success or failure criteria, and a concrete continuation.
\item Once a frontier produces safe novel progress, explore it coherently while observations continue to change.
\item Redirect after escalating danger, loss of a verified return route, repeated parser failures, missing prerequisites, or several actions without novelty.
\item With a strong baseline, place uncertain tests where failure sacrifices as little reliable score as possible.
\item If no strong baseline exists, explore connected novel states systematically using explicit exits, visible objects, and interactions supported by observed changes.
\end{itemize}
\par\vspace{0.2em}
\textbf{Plan beyond discoveries:}
\begin{itemize}[leftmargin=1.25em,itemsep=0pt,topsep=1pt,parsep=0pt,partopsep=0pt]
\item Finding an item, note, NPC, room, or clue is not automatically a scoring success.
\item Distinguish information gain from score gain.
\item Do not prescribe taking or examining objects that were not observed unless clearly labeled as a bounded parser test.
\item After collecting a discovered item, use it only where the log supplies a plausible observed connection; otherwise continue through one selected frontier.
\item If a safe return is verified and further local tests have low value, return by that exact route rather than wandering.
\item When no continuation is verified, choose one testable frontier and define what observations should cause continuation, retreat, or abandonment.
\end{itemize}
\par\vspace{0.2em}
\textbf{Handle terminal actions and budget:}
\begin{itemize}[leftmargin=1.25em,itemsep=0pt,topsep=1pt,parsep=0pt,partopsep=0pt]
\item Identify commands known to win, die, trap, or end useful exploration.
\item Reserve enough turns for a verified finish using observed route cost.
\item Take verified profitable opportunities before finishing when their value exceeds their cost.
\item Do not risk a strong run solely to chase the theoretical maximum.
\item If little budget remains, prioritize an immediate verified scoring or winning action.
\item Do not leave a large remaining budget governed only by "stop," generic exploration, or unspecified improvisation.
\end{itemize}
\par\vspace{0.2em}
\textbf{Adapt to parser feedback:}
\begin{itemize}[leftmargin=1.25em,itemsep=0pt,topsep=1pt,parsep=0pt,partopsep=0pt]
\item Reuse exact nouns, verbs, and prepositions that succeeded in the relevant state.
\item "Not a verb" rejects that verb; do not retry it unchanged.
\item "Can't see any such thing" usually rejects current visibility or availability; check state before changing syntax.
\item "Only understood as far as" confirms only the parsed prefix.
\item Parser paraphrases may reveal the accepted object, argument structure, or a state change.
\item Account for partially successful commands.
\item Limit syntax experiments in hostile, timed, or escalating scenes.
\item Do not prescribe inventory, looking, waiting, help, searching, saving, restoring, or undoing as rituals; they cost turns unless demonstrated useful.
\end{itemize}
\par\vspace{0.2em}
\textbf{Manage inventory and prerequisites:}
\begin{itemize}[leftmargin=1.25em,itemsep=0pt,topsep=1pt,parsep=0pt,partopsep=0pt]
\item Distinguish verified prerequisites, scoring objects, clues, and speculative cargo.
\item Never claim an item is held without acquisition or inventory evidence.
\item Carry an item while its verified or near-term hypothesized purpose remains relevant.
\item Drop items only when capacity pressure or a demonstrated need exists.
\item Count acquisition, dropping, retrieval, and transport costs.
\item Use compound commands only when demonstrated safe and supported.
\end{itemize}
\par\vspace{0.2em}
\textbf{Handle risk from evidence:}
\begin{itemize}[leftmargin=1.25em,itemsep=0pt,topsep=1pt,parsep=0pt,partopsep=0pt]
\item Treat known lethal or trapping transitions as prohibited unless a newly verified prerequisite materially changes the source state.
\item Treat escalating messages as timer evidence. Use a demonstrated solution or a verified exit rather than experimenting broadly.
\item Do not infer safety from absence of warning.
\item Do not claim a plausible tool neutralizes a hazard unless tested.
\item If a hazard warning was ignored, remove nearby optional choices and make the safe command dominant at the exact cue.
\end{itemize}
\par\vspace{0.2em}
\textbf{Write concise, actionable guidance:}
\begin{itemize}[leftmargin=1.25em,itemsep=0pt,topsep=1pt,parsep=0pt,partopsep=0pt]
\item Emphasize the highest-impact adaptation revealed by the latest episode.
\item Preserve the shortest fully verified scoring route, not the shortest imagined route.
\item Separate reliable actions from tests without overstating certainty.
\item Rank alternatives and minimize optional branches.
\item Eliminate obsolete fallbacks, duplicated warnings, generic parser tutorials, and speculative travel.
\item Ensure all recommendations agree on route order, source state, prerequisites, hazards, branch location, and priority.
\item Resolve conflicts in favor of expected score per turn, earlier scoring, baseline preservation, survival or victory, and actor executability.
\end{itemize}
\end{compactpromptbox}

\subsection{WebArena-Lite Meta-Prompt}
\label{app:webarena_prompt}
\begin{compactpromptbox}
You are the META agent in a test-time-learning loop for a browser ACTOR. Read the complete task history and write self-contained guidance for the next episode. Your guidance replaces all prior ACTOR instructions, so preserve every still-useful verified fact needed to finish while removing obsolete procedures, failed speculation, and unnecessary replay.\par
\par\vspace{0.2em}
Optimize for the earliest likely successful episode, not for producing the most detailed plan. The guidance must fit the observed step budget and be robust to the ACTOR's demonstrated behavior.\par
\par\vspace{0.2em}
\textbf{Diagnose adaptation before advising:}
\begin{itemize}[leftmargin=1.25em,itemsep=0pt,topsep=1pt,parsep=0pt,partopsep=0pt]
\item Compare the task goal, observations, actions, errors, final state, evaluator result, prior guidance, and the ACTOR's compliance.
\item Identify the earliest point where the episode diverged from the best supported route.
\item Explain why the prior guidance failed to improve the episode: wrong source, unsupported assumption, low-throughput workflow, ignored instruction, ambiguous action target, tool misuse, repeated setup, excessive branching, incomplete evidence, or omitted final operation.
\item Distinguish a task failure from an adaptation failure. Ask what must change in the next guidance rather than merely what went wrong in the browser.
\item If the ACTOR followed the guidance and it failed, change the strategy or its decomposition. Do not simply restate it more forcefully.
\item Treat generic evaluator rejection narrowly. Detailed evaluator feedback identifying a cause deserves high weight, but still diagnose what caused that condition.
\end{itemize}
\par\vspace{0.2em}
\textbf{Track progress explicitly:}
\begin{itemize}[leftmargin=1.25em,itemsep=0pt,topsep=1pt,parsep=0pt,partopsep=0pt]
\item Decompose the task into independently checkable subgoals.
\item Classify each as verified complete, unresolved, failed, or uncertain.
\item Preserve directly observed facts, exclusions, successful interaction forms, and committed application changes.
\item A failed final result does not erase verified intermediate evidence.
\item Do not make the ACTOR repeat verified work unless later evidence shows it was reset, wrong, or must be rechecked for a concrete reason.
\item Lead with unresolved work. Mention completed work mainly as facts or exclusions needed to prevent replay.
\item Maintain coverage information for searches: what scope has been checked, what remains unchecked, and whether checked records form a useful contiguous range.
\end{itemize}
\par\vspace{0.2em}
\textbf{Enforce the task's proof obligation:}
\begin{itemize}[leftmargin=1.25em,itemsep=0pt,topsep=1pt,parsep=0pt,partopsep=0pt]
\item Determine what evidence is required before an answer or state can be considered correct.
\item For latest, earliest, maximum, or minimum tasks, a matching candidate is not enough: later, earlier, larger, or smaller eligible candidates must be excluded according to the relevant ordering.
\item When records are visibly sorted in the required direction, inspect them contiguously from the decisive end and stop at the first verified match.
\item Do not skip records using weak proxies such as price, size, status, popularity, or apparent likelihood unless the task or visible data establishes that proxy as logically valid.
\item For counts and aggregations, cover the full requested scope and account for pagination, filters, and duplicates.
\item For comparisons, gather equivalent evidence for every required alternative.
\item For state-change tasks, verify the exact target and the committed final state.
\item Never convert a plausible guess into a verified answer.
\end{itemize}
\par\vspace{0.2em}
\textbf{Choose the correct information source:}
\begin{itemize}[leftmargin=1.25em,itemsep=0pt,topsep=1pt,parsep=0pt,partopsep=0pt]
\item Match the task semantics to the source: account history, current application state, record details, search results, settings, map data, repository data, or another visibly relevant source.
\item Product discovery, global search, or general listings do not establish personal history or committed state unless the interface explicitly connects them.
\item Prefer structured records and detail views over indirect inference.
\item Do not invent records, labels, controls, filters, queries, or application behavior.
\item Treat inferred defaults, ordering, identity, and durability as uncertain until observed.
\end{itemize}
\par\vspace{0.2em}
\textbf{Reason about throughput and the step budget:}
\begin{itemize}[leftmargin=1.25em,itemsep=0pt,topsep=1pt,parsep=0pt,partopsep=0pt]
\item Estimate the cost of setup, one iteration of the main loop, recovery, verification, and final submission.
\item Reserve enough budget for the final required operation.
\item Count repeated page-size changes, navigation back to a list, reopening menus, and reloading state as real loop costs.
\item If the projected workflow cannot reach sufficient evidence within one episode, change the route before recommending more iterations.
\item Prefer an observed higher-throughput route such as a larger page, useful sorting, filtering, direct detail navigation, or a consolidated view when available.
\item If a higher-throughput affordance is unobserved, allow only one cheap probe with a clear success condition and a fallback.
\item If no better route is supported, maximize correctness-preserving progress: inspect a contiguous ordered segment rather than scattered "likely" candidates.
\item When a setting repeatedly resets after navigation, include that reset cost in the plan; do not blindly restore it if pagination or another observed route is cheaper.
\item Avoid long candidate lists that cannot be executed within the remaining budget.
\end{itemize}
\par\vspace{0.2em}
\textbf{Adapt after repeated low yield:}
\begin{itemize}[leftmargin=1.25em,itemsep=0pt,topsep=1pt,parsep=0pt,partopsep=0pt]
\item Track whether each episode produced new relevant evidence and how much.
\item New exclusions are progress, but low-rate progress can still make a strategy infeasible.
\item If one episode demonstrates that a serial loop cannot cover the remaining scope within budget, do not continue the same loop unchanged.
\item If two attempts use the same strategy without enough progress to make success likely, retire it unless a specific correctable cause is visible.
\item Change the information source, ordering, decomposition, navigation route, or interaction method rather than adding increasingly speculative guesses.
\item Preserve verified exclusions when changing routes.
\end{itemize}
\par\vspace{0.2em}
\textbf{Learn from action-level evidence:}
\begin{itemize}[leftmargin=1.25em,itemsep=0pt,topsep=1pt,parsep=0pt,partopsep=0pt]
\item Preserve action forms that succeeded, described semantically rather than with transient element IDs.
\item Judge a timeout by the next observation. If navigation or state change completed, treat the action as successful despite the timeout.
\item If an action produced a genuine tool or syntax error and no state change, forbid that form and use the simplest observed working form.
\item After an identical error or unchanged observation repeats, do not retry the same call unchanged.
\item Distinguish automation errors from application rejection; repeated clicking cannot repair malformed input or wrong state.
\item When tool misuse consumes substantial steps, simplify the route and make correct tool use the first adaptation priority.
\end{itemize}
\par\vspace{0.2em}
\textbf{Target controls precisely:}
\begin{itemize}[leftmargin=1.25em,itemsep=0pt,topsep=1pt,parsep=0pt,partopsep=0pt]
\item Before clicking, inspect the semantic role and accessible name of the intended control.
\item When a row, cell, menu, or container contains multiple actions, click the exact child link or button for the intended action.
\item Never recommend clicking a parent row or action cell when it also contains an adjacent destructive, reversible, or state-changing control.
\item Treat an unexpected navigation or mutation as evidence that the wrong child action fired; do not repeat the same target form.
\item If a mistaken side effect is irrelevant to the checker, do not spend scarce steps repairing it. If it affects the task, include the cheapest verified recovery.
\end{itemize}
\par\vspace{0.2em}
Distinguish browser state from committed application state:\par
\begin{itemize}[leftmargin=1.25em,itemsep=0pt,topsep=1pt,parsep=0pt,partopsep=0pt]
\item A fresh episode resets page position, focus, open controls, typed text, and unfinished interactions.
\item A visibly confirmed committed application change is normally durable.
\item Carry confirmed committed changes forward unless later evidence shows reset, rollback, or the wrong target.
\item Never blindly replay toggles, purchases, follows, subscriptions, deletions, reorders, or similar reversible or state-changing actions.
\item If durability is uncertain, advance unresolved work first and recheck only for a concrete reason.
\end{itemize}
\par\vspace{0.2em}
\textbf{Control speculation:}
\begin{itemize}[leftmargin=1.25em,itemsep=0pt,topsep=1pt,parsep=0pt,partopsep=0pt]
\item Label facts as verified only when directly supported by observations or detailed evaluator feedback.
\item Treat hypotheses as hypotheses and use them only to choose a cheap discriminating probe.
\item Do not rank candidates by an unsupported correlation when the task requires exhaustive or ordered evidence.
\item Do not include speculative examples that the ACTOR may copy as answers, targets, or inputs.
\item If guidance depends on an unobserved affordance, state one probe, its observable success condition, and a supported fallback.
\end{itemize}
\par\vspace{0.2em}
\textbf{Ensure completion:}
\begin{itemize}[leftmargin=1.25em,itemsep=0pt,topsep=1pt,parsep=0pt,partopsep=0pt]
\item Identify the exact final operation required by the task or evaluator, including submission to the user when applicable.
\item Put that operation immediately after the minimum sufficient verification; do not postpone it for optional browsing.
\item Preserve the required answer wording or format only when supported by the task, visible interface, or evaluator.
\item Never submit an unverified guess merely because the step limit is near.
\item When sufficient evidence is already present in history, direct immediate submission rather than replaying browser work.
\end{itemize}
\par\vspace{0.2em}
\textbf{Write guidance that is decisive and executable:}
\begin{itemize}[leftmargin=1.25em,itemsep=0pt,topsep=1pt,parsep=0pt,partopsep=0pt]
\item Give one primary plan based on the strongest evidence.
\item Include only a small number of conditionals tied to observations the ACTOR can actually see.
\item Prefer semantic targets, visible labels, record relationships, and success conditions over brittle identifiers.
\item Carry forward task-specific verified facts from the history when useful, but never invent them.
\item Keep the plan short enough that its priority is unmistakable.
\end{itemize}
\end{compactpromptbox}

\subsection{$\tau^2$-bench Meta-Prompt}
\label{app:tau2_prompt}
\begin{compactpromptbox}
You are the META agent in a test-time-learning loop. Read the complete episode log, evaluator evidence, prior guidance, and prior episodes. Write concise, self-contained instructions that help a fresh ACTOR perform the same task better while following the currently visible domain policy.\par
\par\vspace{0.2em}
Diagnose adaptation failure rather than merely describing the latest task error. Determine:\par
\begin{itemize}[leftmargin=1.25em,itemsep=0pt,topsep=1pt,parsep=0pt,partopsep=0pt]
\item which evaluated requirement failed;
\item the earliest behavior that can explain it;
\item whether prior guidance targeted that behavior;
\item whether the ACTOR followed the guidance;
\item whether the targeted check improved;
\item what smallest change to the guidance is justified.
\end{itemize}
\par\vspace{0.2em}
\textbf{Evidence discipline:}
\begin{itemize}[leftmargin=1.25em,itemsep=0pt,topsep=1pt,parsep=0pt,partopsep=0pt]
\item Give highest priority to explicit evaluator evidence: reward components, database checks, action checks, communication checks, termination reason, and tool results.
\item Use the transcript and visible policy to explain evaluator evidence. Do not treat the ACTOR\textquoteright{}s own policy claims as authoritative.
\item Separately assess required reads, required writes, final database state, and communication as confirmed correct, confirmed wrong, or uncertain.
\item Audit every actual mutation in the transcript, including its arguments and side effects.
\item Do not assume an action checker lists or penalizes every extra action. \textquotedblleft{}All expected writes matched\textquotedblright{} can coexist with harmful extra writes and a wrong final database state.
\item Conversely, do not infer which write was extra or missing solely from aggregate counts or reward zero. Verify against the visible policy, request, tool results, and final-state evidence.
\item Preserve checker-confirmed successful outcomes as invariants unless later evidence contradicts them.
\item Identify the earliest divergence sufficient to explain a failed check. Treat later defects as secondary unless they independently affect an evaluated requirement.
\item If the authoritative policy is absent or ambiguous, label the causal interpretation as a hypothesis and instruct the next ACTOR to resolve it from the current policy rather than asserting a guessed rule.
\item If reward, database, and communication checks all pass, stabilize the successful behavior. Do not introduce speculative workflow changes.
\end{itemize}
\par\vspace{0.2em}
\textbf{Causal adaptation:}
\begin{itemize}[leftmargin=1.25em,itemsep=0pt,topsep=1pt,parsep=0pt,partopsep=0pt]
\item Compare episodes while accounting for changes in user wording, supplied inputs, tool results, timing, and termination.
\item For each important prior instruction, determine whether the next ACTOR performed the instructed behavior and whether the corresponding check changed.
\item Credit guidance only when the behavior occurred and the relevant result improved.
\item If prohibited behavior repeats, diagnose guidance-compliance failure rather than repeating the same advice. Shorten the guidance, elevate one visible action gate, and remove distractions.
\item If the ACTOR followed the advice but the check did not improve, revise the underlying hypothesis rather than adding detail.
\item If guidance causes regression on a previously passing check, restore the earlier passing behavior and remove the implicated change unless stronger evidence supports it.
\item Abandon or materially revise a hypothesis that fails repeatedly.
\item Distinguish missing knowledge from failure to execute known steps: use discovery rules for the former and hard pre-action gates for the latter.
\item Prefer one evidence-backed correction to a broad replay plan.
\end{itemize}
\par\vspace{0.2em}
\textbf{Request and state ledger:}
\begin{itemize}[leftmargin=1.25em,itemsep=0pt,topsep=1pt,parsep=0pt,partopsep=0pt]
\item Reconstruct every customer objective:
\begin{itemize}[leftmargin=1.25em,itemsep=0pt,topsep=1pt,parsep=0pt,partopsep=0pt]
\item information requested;
\item state mutation requested;
\item scope and constraints;
\item authorization status;
\item policy prerequisites;
\item required tool inputs;
\item execution result;
\item required outcome communication.
\end{itemize}
\item Track each mutation separately as pending, completed, blocked by policy, failed, or no longer requested.
\item Keep a mutation pending until a write succeeds or evidence establishes that it cannot or should not be performed.
\item Before any response, check whether a pending mutation is executable.
\item A mutation is executable only when:
\begin{itemize}[leftmargin=1.25em,itemsep=0pt,topsep=1pt,parsep=0pt,partopsep=0pt]
\item required authentication and eligibility are established;
\item the visible policy permits the operation itself;
\item the customer has authorized the exact change;
\item every required argument is grounded in customer statements or tool results;
\item any confirmation explicitly required by policy has occurred.
\end{itemize}
\item Distinguish permission to perform an operation from eligibility for a related refund, fee, replacement, credit, or other side effect. Never assume one implies the other.
\item If a mutation is executable, perform it before optional explanations, summaries, offers, or repeated confirmation.
\item If several independent mutations are executable, perform all of them as tool results permit.
\item If a mutation is blocked, identify the exact policy rule or missing prerequisite. Do not perform it merely because the tool accepts the call or the customer insists.
\item If information is missing, ask only for the missing information and bundle related questions when policy permits.
\item Do not invent a confirmation requirement, and do not ask again after valid confirmation has already been supplied.
\item If USER\_STOP occurs on a user turn, assume no later ACTOR turn was available. Recommend moving a write earlier only if it was already authorized, permitted, and fully specified.
\item If termination occurred while required input was missing, improve information gathering rather than inventing the input.
\end{itemize}
\par\vspace{0.2em}
\textbf{Tool and grounding discipline:}
\begin{itemize}[leftmargin=1.25em,itemsep=0pt,topsep=1pt,parsep=0pt,partopsep=0pt]
\item Reconstruct the workflow from current customer statements, visible policy, tool schemas, and tool results.
\item Never guess identifiers, contact details, dates, locations, record references, payment methods, products, or other tool arguments.
\item When an identifier is missing, ask for an accepted identifier or use a policy-supported lookup whose inputs are grounded.
\item Use reads only when they establish authentication, ownership, eligibility, required arguments, requested comparisons, or relevant post-write state.
\item Flag unrelated or speculative reads, but do not let minor inefficiency distract from a failing database or communication requirement.
\item Preserve every customer constraint exactly.
\item For comparative requests, apply all eligibility constraints before optimizing the requested criterion.
\item If grounded results uniquely determine a choice, proceed; if a material tie or preference remains, ask a focused question.
\item Before each write, perform a final gate: requested, authorized, policy-permitted, correctly targeted, and fully grounded.
\item After each write, inspect success or failure, affected records, requested attributes, amounts, status, and side effects.
\item Do not claim success without a successful result.
\item Do not assume one write satisfies another request unless policy or tool output establishes it.
\item Continue remaining authorized operations after partial success when policy permits.
\item Avoid unauthorized writes and disclosure of unrelated account information.
\end{itemize}
\par\vspace{0.2em}
\textbf{Communication discipline:}
\begin{itemize}[leftmargin=1.25em,itemsep=0pt,topsep=1pt,parsep=0pt,partopsep=0pt]
\item Follow the visible domain policy over prior-episode assumptions.
\item Do not convert an ACTOR\textquoteright{}s unsupported explanation or transcript-specific outcome into a general policy rule.
\item Answer every requested informational question using supported facts.
\item Do not state unsupported restrictions, guarantees, timelines, statuses, amounts, refund behavior, delivery behavior, or future outcomes.
\item After writes, report confirmed successes, failures, partial completion, and any material side effects supported by policy or tool results.
\item Explain blocked requests accurately and state what remains unchanged when relevant.
\item Preserve communication items explicitly confirmed by evaluators.
\item Keep prose brief while executable work remains.
\end{itemize}
\par\vspace{0.2em}
\textbf{Writing next-episode guidance:}
\begin{itemize}[leftmargin=1.25em,itemsep=0pt,topsep=1pt,parsep=0pt,partopsep=0pt]
\item Guidance replaces all prior ACTOR instructions, so retain the essential baseline: follow visible policy, authenticate when required, ground all tool arguments, satisfy all requests, execute every authorized and permitted mutation, inspect results, and communicate confirmed outcomes.
\item Put the highest-impact correction first.
\item Express corrections as decision rules with triggers visible to a fresh ACTOR, such as:
\begin{itemize}[leftmargin=1.25em,itemsep=0pt,topsep=1pt,parsep=0pt,partopsep=0pt]
\item a required identifier is missing;
\item authentication or ownership is established;
\item policy eligibility is unresolved;
\item an exact mutation is authorized;
\item all required arguments are known;
\item a write is about to occur;
\item a write succeeded or failed;
\item a response is about to be sent while executable work remains.
\end{itemize}
\item Prefer \textquotedblleft{}verify the operation against the current policy before writing\textquotedblright{} over memorizing an instance-specific eligibility table.
\item Preserve successful behavior explicitly when correcting a regression.
\item After maximal reward, keep the successful causal core stable. Remove only behavior that is clearly irrelevant, unsupported, or risky; do not redesign the workflow for cosmetic improvements.
\item Avoid fixed identifiers, amounts, names, products, routes, dates, records, or memorized tool arguments.
\item Do not prescribe a brittle transcript replay or fixed action sequence when wording, policy, or tool results can change the correct order.
\item Use same-task facts only when indispensable to disambiguate the next attempt, and never present them as reusable domain policy.
\item Clearly distinguish established evidence from hypotheses.
\item Remove unsupported, redundant, contradictory, or distracting advice.
\end{itemize}
\end{compactpromptbox}

\subsection{GEPA Meta-Prompt (Jericho)}
\label{app:gepa_jericho_prompt}
\begin{compactpromptbox}
You are an AI Optimization Assistant.\par
\par\vspace{0.2em}
\textbf{Input Format}\par
The input is a chronological log of a full game episode. Each step in the trajectory follows this structure:
\begin{itemize}[leftmargin=1.25em,itemsep=0pt,topsep=1pt,parsep=0pt,partopsep=0pt]
\item \texttt{[OBS]}: The game state description (what the solver saw).
\item \texttt{[ACTION]}: The command executed by the solver.
\item \texttt{[REWARD]}: The immediate points gained in this step.
\item \texttt{[CUM\_REWARD]}: The total score accumulated so far.
\end{itemize}
\par\vspace{0.2em}
You are a Meta-Agent that reads episodic text-adventure game logs from an Actor Agent and produces \emph{test-time guidance} to help the Actor achieve a higher score in the next episode, under a hard step/turn limit (50). The Actor will replay the same game across episodes; your job is to (1) extract stable game knowledge from prior episodes, (2) diagnose why the last run failed or missed points, and (3) output a concise, actionable next-episode plan (including exact commands when known) that maximizes score while avoiding instant-death traps and wasted steps.\par
\par\vspace{0.2em}
\textbf{INPUT FORMAT YOU WILL SEE}
\begin{itemize}[leftmargin=1.25em,itemsep=0pt,topsep=1pt,parsep=0pt,partopsep=0pt]
\item A multi-episode ``Message History'' containing:
\begin{itemize}[leftmargin=1.25em,itemsep=0pt,topsep=1pt,parsep=0pt,partopsep=0pt]
\item Per-episode logs: [STEP i] with [OBS], [ACTION], [REWARD], [CUM\_REWARD], and a [FINAL\_OBS].
\item After each episode: ``Please reflect\ldots'' plus earlier Meta-Agent feedback (may include errors).
\end{itemize}
\item The final user request: ``Please continue to reflect and provide updated feedback.'' or ``write a new instruction''.
\end{itemize}
\par\vspace{0.2em}
\textbf{YOUR OUTPUT (EACH TIME YOU GIVE GUIDANCE)} (Put it inside \texttt{<learn></learn>})\par
Produce a short strategy update in this structure (use headings):
\begin{enumerate}[leftmargin=1.5em,itemsep=0pt,topsep=1pt,parsep=0pt,partopsep=0pt]
\item ``What we learned (new since last run)''
\item ``Confirmed hazards / do-not-do list''
\item ``Confirmed key mechanics \& exact verbs''
\item ``Next episode plan (step-efficient)''
\begin{itemize}[leftmargin=1.25em,itemsep=0pt,topsep=1pt,parsep=0pt,partopsep=0pt]
\item Provide an ordered command script when possible.
\item Include optional branches only if they fit within the 50-step budget.
\end{itemize}
\item ``Open questions / where the remaining points likely are'' (only if not yet at max score)
\end{enumerate}
\par\vspace{0.2em}
\textbf{GENERAL RULES}
\begin{itemize}[leftmargin=1.25em,itemsep=0pt,topsep=1pt,parsep=0pt,partopsep=0pt]
\item Be action-oriented: prioritize specific next actions, not generic advice.
\item Be step-efficient: explicitly call out detours that cost steps and whether the points are worth it.
\item When the parser is picky, provide exact command phrasing that worked previously.
\item Track known point sources (+10 rooms/items; +100 end capture) and known ``no point'' time sinks.
\item If the last episode ended by death, identify the exact trigger and propose the safe alternative.
\item Do NOT include hidden chain-of-thought; just conclusions and plans.
\end{itemize}
\par\vspace{0.2em}
\textbf{DOMAIN-SPECIFIC GAME KNOWLEDGE TO REMEMBER (CRITICAL; MAY NOT BE IN FUTURE LOGS)}\par
\par\vspace{0.2em}
\textbf{A) Early game / items}
\begin{itemize}[leftmargin=1.25em,itemsep=0pt,topsep=1pt,parsep=0pt,partopsep=0pt]
\item Start: Chief's office. There is a ``piece of white paper'' (take it for +10; reading is optional for points).
\item Chief's office WEST leads to a closet containing a ``small black pistol'' (take pistol for +10). Always grab this early.
\end{itemize}
\par\vspace{0.2em}
\textbf{B) Known instant-death traps}
\begin{itemize}[leftmargin=1.25em,itemsep=0pt,topsep=1pt,parsep=0pt,partopsep=0pt]
\item Restaurant: From the south street area, going NORTH into the ``restraunt'' is INSTANT DEATH (two guys jump you). Avoid approaching the restaurant from that south/street entrance.
\item Dazed man: If you engage/try to pass without proper action, it can kill you. Resolve immediately upon entering his room.
\item Alley EAST: From the alley (after music store), going EAST leads to ``drug house'' = INSTANT DEATH (armed druggies kill you). Avoid.
\item (Unknown) Alley WEST has not been confirmed safe/unsafe; treat as cautious exploration only if step budget allows and you have no better leads.
\end{itemize}
\par\vspace{0.2em}
\textbf{C) Combat / parser mechanics (must be exact)}
\begin{itemize}[leftmargin=1.25em,itemsep=0pt,topsep=1pt,parsep=0pt,partopsep=0pt]
\item ``use gun'' is NOT recognized.
\item ``shoot man'' alone can misparse; the reliable command is:
\begin{itemize}[leftmargin=1.25em,itemsep=0pt,topsep=1pt,parsep=0pt,partopsep=0pt]
\item EXACT: ``shoot man with pistol''
\item Works when you are in the same room as the dazed man (music store encounter room).
\end{itemize}
\item Correct positioning: the dazed man is WEST of the ``back of music store''. You must go WEST first, then immediately shoot.
\end{itemize}
\par\vspace{0.2em}
\textbf{D) Route landmarks that enabled winning (330/360 achieved)}
\begin{itemize}[leftmargin=1.25em,itemsep=0pt,topsep=1pt,parsep=0pt,partopsep=0pt]
\item Street area: From outside west of Chief's office path, STREET has:
\begin{itemize}[leftmargin=1.25em,itemsep=0pt,topsep=1pt,parsep=0pt,partopsep=0pt]
\item EAST $\rightarrow$ Mayor's house (safe, lots of +10 pickups)
\item NORTH $\rightarrow$ restaurant (death; avoid)
\end{itemize}
\item Mayor's house:
\begin{itemize}[leftmargin=1.25em,itemsep=0pt,topsep=1pt,parsep=0pt,partopsep=0pt]
\item Dining room: take ``paper note'' (+10)
\item Living room: take ``wooden wood'' (+10)
\item Upstairs exists; many closets/guest rooms give no points and waste steps.
\item Bedroom (WEST from guard hallway) gives +10 and is optional for progress but needed for max score.
\item Past guard: NORTH to outside (+10)
\end{itemize}
\item Post-house: Outside $\rightarrow$ video store chain $\rightarrow$ outside dead-end $\rightarrow$ music store (+10) $\rightarrow$ back of music store (+10) $\rightarrow$ dazed man fight $\rightarrow$ alley (+10) $\rightarrow$ police station (+10) $\rightarrow$ outside hub with:
\begin{itemize}[leftmargin=1.25em,itemsep=0pt,topsep=1pt,parsep=0pt,partopsep=0pt]
\item WEST $\rightarrow$ ``the wall'' (+10) (safe)
\item EAST $\rightarrow$ ``doughnut king'' (+10) (safe)
\item NORTH $\rightarrow$ ``holiday inn'' (+10), then 15th floor (+10), hallway sequence leads to room \#30 and win (+100)
\end{itemize}
\end{itemize}
\par\vspace{0.2em}
\textbf{E) Step/turn economy facts}
\begin{itemize}[leftmargin=1.25em,itemsep=0pt,topsep=1pt,parsep=0pt,partopsep=0pt]
\item Turn limit is 50; visiting both wall and doughnut king adds 4 moves (go+return each).
\item Holding cells (west/east from police station) have repeatedly given 0 points and waste steps; skip unless pursuing a newly justified lead.
\item Upstairs in mayor's house has many dead-end closets. Use the \emph{minimal} path to get Bedroom (+10) and then Outside (+10). Do not wander.
\end{itemize}
\par\vspace{0.2em}
\textbf{F) Known high-score outcomes}
\begin{itemize}[leftmargin=1.25em,itemsep=0pt,topsep=1pt,parsep=0pt,partopsep=0pt]
\item Winning run: 330/360 in 49 turns is achievable via: pistol $\rightarrow$ mayor's house items $\rightarrow$ bedroom $\rightarrow$ music store (shoot dazed man) $\rightarrow$ alley $\rightarrow$ police station $\rightarrow$ holiday inn $\rightarrow$ room 30.
\item Attempt to add wall + doughnut king without saving steps elsewhere can exceed 50 turns.
\item There remains a missing +10 somewhere (360 max). Candidates previously hypothesized: unexplored branches like Holiday Inn ``east'' from 15th floor, alley west, or other unexplored exits near the video-store outside hub; BUT alley east is confirmed death.
\end{itemize}
\par\vspace{0.2em}
\textbf{WHAT TO DO AFTER EACH NEW EPISODE}
\begin{itemize}[leftmargin=1.25em,itemsep=0pt,topsep=1pt,parsep=0pt,partopsep=0pt]
\item Update the knowledge base with any newly confirmed rooms, deaths, verbs, and point awards.
\item If the Actor dies, identify the precise action that triggered death and add it to hazards.
\item Propose the next run's command script that:
\begin{enumerate}[leftmargin=1.5em,itemsep=0pt,topsep=1pt,parsep=0pt,partopsep=0pt]
\item guarantees survival (avoid known deaths),
\item reaches win condition,
\item adds extra +10 locations only if step budget permits.
\end{enumerate}
\end{itemize}
\par\vspace{0.2em}
\textbf{STYLE CONSTRAINTS}
\begin{itemize}[leftmargin=1.25em,itemsep=0pt,topsep=1pt,parsep=0pt,partopsep=0pt]
\item Keep guidance compact and executable.
\item Prefer a numbered command list for the next episode.
\item Use bold for ``DO NOT'' hazards and for exact command syntax that must be typed.
\end{itemize}
\par\vspace{0.2em}
\textbf{Output Format:}\par
\texttt{<think>reasoning\ldots</think>}\par
\texttt{<learn>Your derived useful feedback. This will be used as the new system prompt for the actor agent</learn>}\par
\par\vspace{0.2em}
If no useful feedback can be derived, output an empty \texttt{<learn></learn>}.
\end{compactpromptbox}

\section{Case Studies}
\label{app:case_studies}

This section walks through three TTL sessions, one per benchmark, each run with the optimized meta-prompt (Appendix~\ref{app:optimized_prompts}). For each session, we show what the meta-agent wrote between episodes and how the actor's behavior changed as a result. We also point out where the three strategies from Section~\ref{sec:rq3} appear: meta-level credit assignment, verified knowledge retention, and bounded exploration.

\paragraph{Setup.} In a TTL session, the actor attempts the same task over several episodes, and the environment is reset at the start of each episode. Within an episode, the actor acts based only on its system prompt and current observation. After each episode, the meta-agent, following the meta-prompt, reads the history of previous episodes and writes new guidance for the actor. This guidance becomes the actor's entire system prompt for the next episode, so it is the only information carried from one episode to the next. Text in quotation marks is copied exactly from the environment, the meta-agent's output, or the meta-prompt.

\subsection{Jericho: \textit{Balances} (OOD)}
\label{app:case_jericho}

\paragraph{Task.} Jericho games are text adventures: the actor types short commands (e.g., \texttt{take cube}), the game replies with text, and some actions earn points. Each episode allows at most 50 steps, and the session has 6 episodes. \textit{Balances} is an OOD game, never seen during meta-training. The player gains spells by copying scrolls into a spell book, and must memorize a spell with \texttt{learn} before each use. Here, the guidance is a numbered list of commands for the actor to type from the start of the game.

\begin{center}
\small
\begin{tabular}{lcccccc}
\toprule
Episode & 1 & 2 & 3 & 4 & 5 & 6 \\
\midrule
Score & 10 & 15 & 15 & 15 & 20 & 25 \\
\bottomrule
\end{tabular}
\end{center}

\paragraph{After episode 1.} Starting from a generic prompt, the actor earned 5 points twice (\texttt{examine furniture} and \texttt{examine oats}) and then wandered. The new command list (1) starts with these two commands; (2) keeps a rule confirmed by the game, ``Learn a spell before casting it''; and (3) asks the actor to try one idea whose outcome is not yet known, to check a guess: the actor had copied the spell bozbar, which ``causes an animal to sprout wings'', so cast it on a nearby horse and ride it, trying \texttt{mount horse} if \texttt{ride horse} is not accepted. In episode 2, the horse flew the actor across a chasm to a cave, earning 5 more points.

\paragraph{Episodes 2--4: the scales.} In the cave, taking coins off a pair of scales traps the actor inside (``something bars your way \ldots\ you were trying to steal its coins!''). The meta-agent's first guess, adding the actor's silver coin to the left side, failed in episode 3. Its second guess was to cast the spell urbzig on the scales, which the guidance presented as something to try rather than a known solution; the game rejected it in episode 4. Since the actor had followed the guidance exactly, the meta-agent concluded that the guess itself was wrong and dropped it, instead of suggesting different wording:
\begin{quote}
``Urbzig is not for the scales. Its description (`turn a dangerous object into a harmless one') clearly matches the hissing snake at the chasm, not the scales.''
\end{quote}
It then worked out the answer from the contents of the scales recorded in the history: copying the scroll on the left side had made that side lighter, and adding the silver coin had made it too heavy, so the silver coin should \emph{replace} the bronze coin. Because every command list still began with the commands that earned the first 15 points, the two wrong guesses cost no points.

\paragraph{Episodes 5--6.} Replacing the coin worked (``the scales come into balance''). The actor then cast urbzig on the snake (``the snake is replaced by a clump of daffodils'') and took the cube it guarded (score 20). The next guidance kept both findings together with where they apply, and suggested one more thing to try, with a backup plan in case it failed:
\begin{quote}
``A bounded, low-risk test is to cast Bozbar on the tortoise, mount it, and attempt to go up at the temple. If that fails quickly, shift to \ldots\ the footpath north from Cave Mouth.''
\end{quote}
The tortoise flew away instead, but left behind a scroll worth 5 points (score 25), and the actor then moved on to the footpath as planned.

\paragraph{Strategies shown.}
\begin{itemize}[leftmargin=1.25em,itemsep=1pt,topsep=2pt]
\item \textbf{Bounded exploration.} Each command list repeats the commands known to earn points before adding one new thing to try, so a failed attempt never costs points already earned.
\item \textbf{Verified knowledge retention.} Guesses were given to the actor as things to try, not as facts. Only findings confirmed by the game were kept, each with the situation it applies to.
\item \textbf{Meta-level credit assignment.} In episode 4, the actor followed the guidance and it still failed, so the meta-agent attributed the failure to the guidance and replaced the idea instead of repeating it.
\end{itemize}

\subsection{WebArena-Lite: Task 78 (ID, Shopping)}
\label{app:case_webarena}

\paragraph{Task.} The actor answers questions on a shopping website by clicking and typing, with at most 10 actions per episode and 3 episodes per session. After each episode, the meta-agent learns only whether the answer was correct, plus a short reason. The task is:
\begin{quote}
``How much did I spend on shopping at One Stop Market on November 2022? They gave me a 20\% discount on the total amount for orders exceeding \$200 in cash.''
\end{quote}
The actor must add up all November 2022 orders, taking 20\% off any cash order over \$200. The three episodes failed, failed, and passed.

\paragraph{Episode 1: only one order counted.} The actor found the order list but answered with the discounted amount of a single order (\$174.54), leaving out two other November orders shown on the same page. The meta-agent identified the cause:
\begin{quote}
``they stopped after inspecting a single order \ldots\ while the task asks for total spending in November 2022.''
\end{quote}
The new guidance told the actor to collect every November order before answering. It also copied in the facts already seen: the list is sorted newest first, all November orders fit on the first page at 50 orders per page, the three amounts, and only one order exceeds \$200.

\paragraph{Episode 2: no answer sent.} Several clicks reported a timeout error even though the page had changed, so the actor repeated the same steps and ran out of actions before answering. The meta-agent made two changes: treat a timed-out click as successful if the page changed, and, since all numbers were already known, write out the calculation ($218.17 \times 0.8 + 133.07 + 51.94 = 359.55$) and submit right after one check.

\paragraph{Episode 3: pass.} The actor checked the order and sent ``\$359.55'' using 5 actions.

\paragraph{Strategies shown.}
\begin{itemize}[leftmargin=1.25em,itemsep=1pt,topsep=2pt]
\item \textbf{Meta-level credit assignment.} The two failures had different causes (missing orders, then no answer sent), and the meta-agent fixed each one directly instead of giving general advice.
\item \textbf{Verified knowledge retention.} Facts seen on the page were carried forward, so the actor did not search again after failed episodes. The meta-prompt states: ``A failed final result does not erase verified intermediate evidence.''
\end{itemize}

\subsection{$\tau^2$-bench: Airline Task 32 (ID)}
\label{app:case_tau2}

\paragraph{Task.} The actor is an airline customer-service agent. It chats with a simulated user and, following the airline's policy, uses tools to look up and change bookings. An episode scores 1 only if (a) the database ends in the correct state and (b) the user was told the required information; otherwise it scores 0. The simulated user can end the conversation at any time, and the session has 3 episodes. Here, the user wants to move a basic economy booking to the nonstop flight HAT041. Since basic economy cannot be changed, this takes two changes: an upgrade to economy, then the flight change. The three episodes scored 0, 1, and 1.

\paragraph{Episode 1: an approved change was never made.} After the upgrade, the user said:
\begin{quote}
``Please go ahead and switch us to flight HAT041. You can put the refund back onto my credit card ending in 1777.''
\end{quote}
Instead of making the change, the actor asked ``Do you wish to proceed with this change?'' The user said yes and ended the conversation, so the change was never made and check (a) failed. The meta-agent traced the failure to this point:
\begin{quote}
``The agent asked for redundant confirmation after the user had already given complete authorization and all required inputs.''
\end{quote}
It also noted that the previous guidance never covered this situation, so the failure came from missing guidance rather than the actor ignoring it. It added one rule and kept the upgrade steps unchanged: once the user has clearly approved a specific change that the policy allows, make it right away instead of asking again.

\paragraph{Episodes 2--3: the rule works.} The actor made the change as its next action, and both checks passed. Before keeping the rule, the meta-agent checked that the actor had actually followed it and that check (a) now passed. It then removed only two things: an unsupported promise that the refund would arrive ``in 5 to 7 business days'', and an unneeded lookup of the airport list. Episode 3 passed again.

\paragraph{Strategies shown.}
\begin{itemize}[leftmargin=1.25em,itemsep=1pt,topsep=2pt]
\item \textbf{Meta-level credit assignment.} Before adding the rule, the meta-agent checked whether the previous guidance covered the failure; before keeping it, it checked that the actor followed it and that it helped. The meta-prompt states: ``Credit guidance only when the behavior occurred and the relevant result improved.''
\item \textbf{Verified knowledge retention.} Behavior confirmed by the checks was kept unchanged, and only unsupported or unneeded parts were removed.
\end{itemize}
Note that the policy requires the user's explicit approval before any change, and here the user had already given it. The learned rule is to not ask twice, not to skip approval.

\section{Algorithm descriptions}
\label{app:algo_descriptions}
\begin{algorithm}[H]
\caption{\textsc{RunTTL}$(\phi, g)$}
\label{alg:run_ttl}
\small
\begin{algorithmic}[1]
\Require Meta-prompt $\phi$; task $g$; initial actor prompt $\rho_1$; episode budget $K$
\For{$k = 1, \ldots, K$}
    \State $\tau_k \leftarrow \textsc{RunActor}(g, \rho_k)$
    \If{$k < K$}
        \State $\rho_{k+1} \leftarrow \textsc{Adapt}(\phi, \{\tau_1, \ldots, \tau_k\})$
    \EndIf
\EndFor
\State \textbf{return} $\xi_g^{\phi} \leftarrow (\tau_1, \ldots, \tau_K)$
\end{algorithmic}
\end{algorithm}

\begin{algorithm}[H]
  \caption{Reflective Meta-Learning of the Self-Improvement Policy}
  \label{alg:meta_training}
  \small
  \begin{algorithmic}[1]
  \Require Seed meta-prompt $\phi_0$; training tasks $\mathcal{D}_{\mathrm{train}}$;
  validation tasks $\mathcal{D}_{\mathrm{val}}$; iteration budget $T$;
  acceptance margin $\delta$; episode-0 baseline results
  \State $\phi^\star \leftarrow \phi_0$
  \State Initialize per-validation-task score histories
  $\mathcal{H} = \{\mathcal{H}_h = \emptyset : h \in \mathcal{D}_{\mathrm{val}}\}$

  \For{$t = 1, \ldots, T$}
      \State Sample $g \sim \mathcal{D}_{\mathrm{train}}$

      \State $\xi_g \leftarrow
      \textsc{RunTTL}(\phi^\star, g)$
      \algcomment{run the current expert on one proposal task}

      \State $F_g \leftarrow \textsc{ExtractFacts}(\phi^\star, \xi_g)$
      \algcomment{attribute task-specific facts from the complete TTL session}

      \State $\tilde{\phi} \leftarrow
      \textsc{Propose}(\phi^\star, \xi_g, F_g)$
      \algcomment{reflect on the adaptation failure and propose a task-agnostic strategy}

      \State $\mathcal{V}_t \leftarrow \emptyset$
      \For{$h \in \mathcal{D}_{\mathrm{val}}$}
          \State $\xi_h^\star \leftarrow
          \textsc{RunTTL}(\phi^\star, h)$
          \State $\tilde{\xi}_h \leftarrow
          \textsc{RunTTL}(\tilde{\phi}, h)$
          \algcomment{global validation of current and proposed meta-prompts}

            \State $s_h^\star \leftarrow \text{W-AUC}(\xi_h^\star)$
            \State $\tilde{s}_h \leftarrow \text{W-AUC}(\tilde{\xi}_h)$
             \algcomment{calculate the W-AUC scores of the resulting TTL sessions}
            \State $\mathcal{V}_t \leftarrow \mathcal{V}_t \cup \{h\}$
         
      \EndFor

      \For{$h \in \mathcal{V}_t$}
          \State $(z_h^\star,\tilde{z}_h) \leftarrow
          \textsc{Normalize}
          (s_h^\star,\tilde{s}_h;\mathcal{H}_h)$
          \algcomment{normalize both scores against the same per-task pool}
      \EndFor

      \State $\bar{z}^\star \leftarrow
      \frac{1}{|\mathcal{V}_t|}
      \sum_{h \in \mathcal{V}_t} z_h^\star$
      \State $\bar{\tilde{z}} \leftarrow
      \frac{1}{|\mathcal{V}_t|}
      \sum_{h \in \mathcal{V}_t} \tilde{z}_h$
      \algcomment{calculate the average normalized score for both meta-prompts}

      \For{$h \in \mathcal{V}_t$}
          \State $\mathcal{H}_h \leftarrow
          \mathcal{H}_h \cup \{s_h^\star,\tilde{s}_h\}$
      \EndFor

      \If{$\bar{\tilde{z}} > \bar{z}^\star$}
          \State $\phi^\star \leftarrow \tilde{\phi}$
          \algcomment{replace the current expert}
      \EndIf
  \EndFor

  \State \textbf{return} $\phi^\star$
\end{algorithmic}
\end{algorithm}

\section{Evaluation with Different Model Backbones}
\label{app:backbones}

To show the generality of our framework across different model backbones, we provide additional results on Jericho and WebArena-Lite with GLM-5.3-Flash as the meta-agent. The meta-prompt used for evaluation is independently meta-trained with this backbone. Baselines that use an LM to update the actor's context also use GLM-5.3-Flash, and the actor remains Gemini-3-Flash. Tables~\ref{tab:backbones_jericho} and~\ref{tab:backbones_webarena} report average score and W-AUC, each averaged over the ID or OOD tasks. \modelname achieves the best result in every column on both benchmarks, consistent with the main results. 

Below is an excerpt from the meta-prompt learned with GLM-5.3-Flash on Jericho:
\begin{quote}
\small
``Determine whether the last strategy failed because its hypothesis was poor, its execution was infeasible, the actor did not comply, or the post-script policy was too vague.''

``Distinguish internally: Verified facts: directly demonstrated by observations or action results. \ldots\ Hypotheses: plausible but untested. Never present a hypothesis as fact or as the unique solution.''

``Select one primary experiment for the next episode. Add at most one simple fallback, triggered by a specific observable failure of the primary plan.''
\end{quote}
These instructions align with the identified three common strategies discussed in Section~\ref{sec:rq3}: meta-level credit assignment, verified knowledge retention, and bounded exploration. Similar strategies emerge with a different meta-agent backbone, suggesting that they are not specific to GPT-5.

\input{tables/backbones_jericho}
\input{tables/backbones_webarena}

\section{Implementation Details of Parameter-Based TTL Methods}
\label{app:parameter_ttl}

\paragraph{Setup.} All methods use Qwen3-8B as the actor backbone, TTL sessions of $K=3$ episodes, and each episode is contains at most 30 actor steps. We compare \modelname with three parameter-based TTL methods, which update the actor's weights between episodes. For \modelname, the meta-agent also uses Qwen3-8B, matching the actor backbone, and runs the meta-prompt learned in our main experiments.

\paragraph{Online SFT.} After each episode, we convert the trajectory into a dataset of (state, action) pairs, and the actor is fine-tuned on this dataset with a standard supervised fine-tuning objective before the next episode. To avoid imitating harmful behavior, we filter the dataset prior to finetuning: we keep only (i) episodes whose score is at least the best score seen so far on that game, and, within each kept episode, (ii) state-action pairs followed by non-zero future reward. Each update runs one epoch over the filtered data (batch size 1, one gradient step per sample) with a learning rate of $1\times10^{-5}$.

\paragraph{Online GRPO.} We update the model between each episode during one TTL session. After each episode, the trajectory and its rewards are used to compute a GRPO policy gradient \citep{shao2024deepseekmathpushinglimitsmathematical}, and the actor's weights are updated before the next episode. Each update uses a group size of 8 and a learning rate of $1\times10^{-6}$.

\paragraph{SFT+MAML.} This baseline meta-learns an initialization for Online SFT. On the ID games, a first-order MAML outer loop, implemented with Reptile \citep{finn2017modelagnosticmetalearningfastadaptation,nichol2018firstordermetalearningalgorithms}, learns the initialization $\theta_0$ of a LoRA adapter \citep{hu2021loralowrankadaptationlarge} for the Online SFT inner loop. At test time, the actor starts from $\theta_0$ and adapts with Online SFT. The LoRA adapter trained using 15 meta-steps on 2xH200 GPUs, and the 8B base model is kept frozen. We use first-order Reptile to avoid the heavy computational cost of second-order MAML, which requires computing second-order gradients through the inner loop of an 8B model.

\end{document}

%% file: tables/jericho.tex
\begin{table*}[t]
\centering
\small
\setlength{\tabcolsep}{5.0pt}
\caption{\textbf{Jericho Results.} Top: ID games used for meta-training; bottom: OOD games unseen during meta-training. All methods use a frozen Gemini-3-Flash actor. Methods requiring a separate adaptation LLM use GPT-5. Each result is averaged over five repeated runs. Best results are in \textbf{bold} and second-best results are \underline{underlined}. Green subscripts show the relative improvement of \modelname's average over the strongest baseline.}
\label{tab:jericho_id}\label{tab:jericho_ood}
\begin{tabular}{@{}l*{8}{c}@{}}
\toprule
\textbf{Method} & \multicolumn{4}{c}{\textbf{Avg. Score $\uparrow$}}
& \multicolumn{4}{c}{\textbf{W-AUC $\uparrow$}} \\
\cmidrule(lr){2-5} \cmidrule(l){6-9}
\rowcolor{blockgray}
\multicolumn{1}{@{}>{\columncolor{blockgray}[0pt][\tabcolsep]}l}{\rule[-0.7ex]{0pt}{2.8ex}\textit{\textbf{ID}}}
& Detective & Zork 1 & Temple & Avg.
& Detective & Zork 1 & Temple & \multicolumn{1}{>{\columncolor{blockgray}[\tabcolsep][0pt]}c@{}}{Avg.} \\
\midrule
\multicolumn{9}{@{}l}{\textbf{\textit{Non-Learning Baseline}}} \\
Static & 122.00 & 39.83 & 5.00 & 55.61 & 0.328 & 0.114 & 0.143 & 0.195 \\
\midrule
\multicolumn{9}{@{}l}{\textbf{\textit{Reflection- \& Memory-Based Methods}}} \\
Reflexion & 101.33 & 41.47 & 4.03 & 48.94 & 0.288 & 0.120 & 0.113 & 0.173 \\
Memory Agent & 102.00 & 44.57 & 5.00 & 50.52 & 0.268 & 0.132 & 0.143 & 0.181 \\
AWM & 148.33 & 44.93 & 5.80 & 66.36 & 0.441 & 0.132 & 0.172 & 0.249 \\
\midrule
\multicolumn{9}{@{}l}{\textbf{\textit{Prompt \& Configuration Optimization Methods}}} \\
TextGrad & 88.67 & 40.40 & 5.27 & 44.78 & 0.218 & 0.108 & 0.140 & 0.156 \\
EvoPrompt & 108.33 & 45.27 & 5.77 & 53.12 & 0.299 & 0.128 & 0.161 & 0.196 \\
EvoTest & \underline{205.33} & \underline{48.03} & \underline{7.03} & \underline{86.80} & \underline{0.628} & \underline{0.139} & \underline{0.202} & \underline{0.323} \\
Naive Meta-Agent & 114.00 & 44.30 & 6.73 & 55.01 & 0.327 & 0.129 & 0.189 & 0.215 \\
\midrule
\rowcolor{oursrow}
\multicolumn{1}{@{}>{\columncolor{oursrow}[0pt][\tabcolsep]}l}{\modelname}
& \textbf{252.67} & \textbf{50.33} & \textbf{8.23} & \textbf{103.74}\gain{20\%} & \textbf{0.790} & \textbf{0.150} & \textbf{0.248}
& \multicolumn{1}{>{\columncolor{oursrow}[\tabcolsep][0pt]}c@{}}{\textbf{0.396}\gain{23\%}} \\
\midrule[\heavyrulewidth]
\rowcolor{blockgray}
\multicolumn{1}{@{}>{\columncolor{blockgray}[0pt][\tabcolsep]}l}{\rule[-0.7ex]{0pt}{2.8ex}\textit{\textbf{OOD}}}
& Balances & Library & Zork 3 & Avg.
& Balances & Library & Zork 3 & \multicolumn{1}{>{\columncolor{blockgray}[\tabcolsep][0pt]}c@{}}{Avg.} \\
\midrule
\multicolumn{9}{@{}l}{\textbf{\textit{Non-Learning Baseline}}} \\
Static & 7.33 & 5.50 & 2.50 & 5.11 & 0.142 & 0.184 & 0.356 & 0.228 \\
\midrule
\multicolumn{9}{@{}l}{\textbf{\textit{Reflection- \& Memory-Based Methods}}} \\
Reflexion & 7.17 & 8.13 & 1.97 & 5.76 & 0.141 & 0.313 & 0.268 & 0.241 \\
Memory Agent & 8.50 & 7.47 & 1.77 & 5.91 & 0.172 & 0.283 & 0.219 & 0.224 \\
AWM & 8.50 & 8.87 & \underline{2.67} & 6.68 & 0.170 & 0.329 & 0.381 & 0.293 \\
\midrule
\multicolumn{9}{@{}l}{\textbf{\textit{Prompt \& Configuration Optimization Methods}}} \\
TextGrad & 8.67 & 10.87 & 1.83 & 7.12 & 0.164 & 0.347 & 0.244 & 0.252 \\
EvoPrompt & 9.33 & \underline{12.17} & 2.47 & \underline{7.99} & 0.184 & 0.396 & 0.351 & 0.310 \\
EvoTest & 9.67 & 10.30 & \textbf{2.87} & 7.61 & 0.197 & 0.355 & \underline{0.423} & \underline{0.325} \\
Naive Meta-Agent & \underline{10.00} & 11.97 & 1.73 & 7.90 & \underline{0.200} & \underline{0.409} & 0.248 & 0.286 \\
\midrule
\rowcolor{oursrow}
\multicolumn{1}{@{}>{\columncolor{oursrow}[0pt][\tabcolsep]}l}{\modelname}
& \textbf{13.67} & \textbf{15.30} & \textbf{2.87} & \textbf{10.61}\gain{33\%} & \textbf{0.292} & \textbf{0.522} & \textbf{0.427}
& \multicolumn{1}{>{\columncolor{oursrow}[\tabcolsep][0pt]}c@{}}{\textbf{0.414}\gain{27\%}} \\
\bottomrule
\end{tabular}
\end{table*}

%% file: tables/webarena.tex
\begin{table*}[t]
\centering
\small
\caption{\textbf{WebArena-Lite Results.} Top: ID domains used for meta-training; bottom: OOD domains unseen during meta-training. All methods use a frozen Gemini-3-Flash actor. Methods requiring a separate adaptation LLM use GPT-5. Each result is averaged over three repeated runs. Best results are in \textbf{bold} and second-best results are \underline{underlined}. Green subscripts show the relative improvement of \modelname's average over the strongest baseline.}
\label{tab:webarena_id}\label{tab:webarena_ood}
\begin{tabular}{@{}l*{8}{c}@{}}
\toprule
\textbf{Method} & \multicolumn{4}{c}{\textbf{Avg. Score $\uparrow$}}
& \multicolumn{4}{c}{\textbf{W-AUC $\uparrow$}} \\
\cmidrule(lr){2-5} \cmidrule(l){6-9}
\rowcolor{blockgray}
\multicolumn{1}{@{}>{\columncolor{blockgray}[0pt][\tabcolsep]}l}{\rule[-0.7ex]{0pt}{2.8ex}\textit{\textbf{ID}}}
& GitLab & Map & Shopping & Avg.
& GitLab & Map & Shopping & \multicolumn{1}{>{\columncolor{blockgray}[\tabcolsep][0pt]}c@{}}{Avg.} \\
\midrule
\multicolumn{9}{@{}l}{\textbf{\textit{Non-Learning Baseline}}} \\
Static & 0.811 & 0.322 & 0.533 & 0.556 & 0.817 & 0.328 & 0.544 & 0.563 \\
\midrule
\multicolumn{9}{@{}l}{\textbf{\textit{Reflection- \& Memory-Based Methods}}} \\
Reflexion & 0.822 & 0.344 & 0.511 & 0.559 & 0.856 & 0.356 & 0.528 & 0.580 \\
Memory Agent & 0.822 & \underline{0.378} & \underline{0.611} & \underline{0.604} & 0.828 & 0.383 & \underline{0.644} & \underline{0.619} \\
AWM & \underline{0.867} & 0.333 & 0.433 & 0.544 & \textbf{0.883} & 0.344 & 0.444 & 0.557 \\
\midrule
\multicolumn{9}{@{}l}{\textbf{\textit{Prompt \& Configuration Optimization Methods}}} \\
TextGrad & 0.800 & 0.356 & 0.567 & 0.574 & 0.800 & 0.378 & 0.589 & 0.589 \\
EvoPrompt & 0.833 & 0.311 & 0.522 & 0.556 & 0.844 & 0.317 & 0.550 & 0.570 \\
EvoTest & \underline{0.867} & \underline{0.378} & 0.522 & 0.589 & \underline{0.867} & \underline{0.394} & 0.539 & 0.600 \\
Naive Meta-Agent & 0.778 & 0.356 & 0.500 & 0.545 & 0.806 & 0.383 & 0.511 & 0.567 \\
\midrule
\rowcolor{oursrow}
\multicolumn{1}{@{}>{\columncolor{oursrow}[0pt][\tabcolsep]}l}{\modelname}
& \textbf{0.878} & \textbf{0.444} & \textbf{0.644} & \textbf{0.656}\gain{9\%} & \textbf{0.883} & \textbf{0.483} & \textbf{0.711}
& \multicolumn{1}{>{\columncolor{oursrow}[\tabcolsep][0pt]}c@{}}{\textbf{0.693}\gain{12\%}} \\
\midrule[\heavyrulewidth]
\rowcolor{blockgray}
\multicolumn{1}{@{}>{\columncolor{blockgray}[0pt][\tabcolsep]}l}{\rule[-0.7ex]{0pt}{2.8ex}\textit{\textbf{OOD}}}
& Reddit & \multicolumn{2}{c}{Shopping Admin} & Avg.
& Reddit & \multicolumn{2}{c}{Shopping Admin} & \multicolumn{1}{>{\columncolor{blockgray}[\tabcolsep][0pt]}c@{}}{Avg.} \\
\midrule
\multicolumn{9}{@{}l}{\textbf{\textit{Non-Learning Baseline}}} \\
Static & 0.655 & \multicolumn{2}{c}{0.716} & 0.685 & 0.664 & \multicolumn{2}{c}{0.724} & 0.694 \\
\midrule
\multicolumn{9}{@{}l}{\textbf{\textit{Reflection- \& Memory-Based Methods}}} \\
Reflexion & 0.678 & \multicolumn{2}{c}{0.733} & 0.706 & 0.684 & \multicolumn{2}{c}{0.744} & 0.714 \\
Memory Agent & 0.643 & \multicolumn{2}{c}{\underline{0.738}} & 0.691 & 0.646 & \multicolumn{2}{c}{0.742} & 0.694 \\
AWM & 0.696 & \multicolumn{2}{c}{0.720} & 0.708 & 0.705 & \multicolumn{2}{c}{0.731} & 0.718 \\
\midrule
\multicolumn{9}{@{}l}{\textbf{\textit{Prompt \& Configuration Optimization Methods}}} \\
TextGrad & 0.702 & \multicolumn{2}{c}{0.729} & \underline{0.715} & \underline{0.719} & \multicolumn{2}{c}{0.737} & \underline{0.728} \\
EvoPrompt & 0.684 & \multicolumn{2}{c}{\underline{0.738}} & 0.711 & 0.684 & \multicolumn{2}{c}{0.742} & 0.713 \\
EvoTest & 0.678 & \multicolumn{2}{c}{0.733} & 0.706 & 0.690 & \multicolumn{2}{c}{\underline{0.749}} & 0.719 \\
Naive Meta-Agent & \underline{0.708} & \multicolumn{2}{c}{0.702} & 0.705 & \underline{0.719} & \multicolumn{2}{c}{0.724} & 0.722 \\
\midrule
\rowcolor{oursrow}
\multicolumn{1}{@{}>{\columncolor{oursrow}[0pt][\tabcolsep]}l}{\modelname}
& \textbf{0.754} & \multicolumn{2}{c}{\textbf{0.747}} & \textbf{0.751}\gain{5\%} & \textbf{0.772} & \multicolumn{2}{c}{\textbf{0.760}}
& \multicolumn{1}{>{\columncolor{oursrow}[\tabcolsep][0pt]}c@{}}{\textbf{0.766}\gain{5\%}} \\
\bottomrule
\end{tabular}
\end{table*}

%% file: tables/tau2bench.tex
\begin{table}[t]
\centering
\small
\caption{\textbf{$\tau^2$-bench Results.} ID results on Airline and Retail, and OOD results on Telecom with the meta-prompt frozen. All methods use a frozen Gemini-3-Flash actor. Methods requiring a separate adaptation LLM use GPT-5. Best results are in \textbf{bold} and second-best results are \underline{underlined}. Green subscripts show the relative W-AUC improvement of \modelname over the strongest baseline.}
\label{tab:tau2bench}
\begin{tabular}{lcccccc}
\toprule
& \multicolumn{3}{c}{\textbf{Avg. Score} $\uparrow$} & \multicolumn{3}{c}{\textbf{W-AUC} $\uparrow$} \\
\cmidrule(lr){2-4} \cmidrule(lr){5-7}
\textbf{Method} & Airline & Retail & Telecom (OOD) & Airline & Retail & Telecom (OOD) \\
\midrule
\multicolumn{7}{l}{\textbf{\textit{Non-Learning Baseline}}} \\
Static & 0.400 & 0.450 & 0.281 & 0.450 & 0.425 & 0.272 \\
\midrule
\multicolumn{7}{l}{\textbf{\textit{Reflection- \& Memory-Based Methods}}} \\
Reflexion & 0.567 & \textbf{0.633} & \underline{0.517} & 0.617 & 0.600 & \underline{0.525} \\
Memory Agent & 0.500 & \underline{0.617} & 0.350 & 0.483 & \underline{0.633} & 0.392 \\
AWM & 0.467 & 0.567 & 0.450 & 0.483 & 0.542 & 0.467 \\
\midrule
\multicolumn{7}{l}{\textbf{\textit{Prompt \& Configuration Optimization Methods}}} \\
TextGrad & \underline{0.600} & 0.567 & 0.354 & \underline{0.650} & 0.608 & 0.365 \\
EvoPrompt & 0.467 & 0.550 & 0.278 & 0.483 & 0.550 & 0.269 \\
EvoTest & 0.567 & 0.583 & 0.370 & 0.567 & 0.617 & 0.417 \\
Naive Meta-Agent & \underline{0.600} & 0.533 & 0.400 & 0.633 & 0.558 & 0.433 \\
\midrule
\rowcolor{oursrow}
\modelname & \textbf{0.667} & \textbf{0.633} & \textbf{0.533} & \textbf{0.717}\gain{10\%} & \textbf{0.700}\gain{11\%} & \textbf{0.583}\gain{11\%} \\
\bottomrule
\end{tabular}
\end{table}

%% file: tables/parameter_ttl.tex
\begin{table*}[!t]
    \centering
    \small
    \caption{\textbf{Parameter-updating comparison on Jericho (W-AUC).} Online SFT, Online GRPO, and SFT+MAML update a single Qwen3-8B actor. \modelname uses a Qwen3-8B actor with a separate Qwen3-8B meta-agent that runs the learned meta-policy from the main experiment.}
    \label{tab:parameter_ttl}
    \begin{tabularx}{\textwidth}{@{}l*{8}{Y}@{}}
        \toprule
        \multirow{2}{*}{\textbf{Method}} & \multicolumn{4}{c}{\textbf{ID}} & \multicolumn{4}{c}{\textbf{OOD}} \\
        \cmidrule(lr){2-5} \cmidrule(lr){6-9}
        & \textbf{Detective} & \textbf{Zork~1} & \textbf{Temple} & \textbf{Avg.}
        & \textbf{Balances} & \textbf{Library} & \textbf{Zork~3} & \textbf{Avg.} \\
        \midrule
        Static      & 0.162 & 0.019 & 0.071 & 0.084 & 0.114 & 0.000 & 0.167 & 0.094 \\
        Online GRPO & 0.139 & 0.014 & 0.200 & 0.118 & \textbf{0.147} & 0.194 & 0.000 & 0.114 \\
        Online SFT  & 0.185 & 0.005 & 0.157 & 0.116 & 0.098 & 0.083 & 0.095 & 0.092 \\
        SFT+MAML    & 0.185 & 0.038 & 0.143 & 0.122 & 0.065 & 0.000 & \textbf{0.405} & 0.157 \\
        \rowcolor{oursrow}
        \multicolumn{1}{@{}>{\columncolor{oursrow}[0pt][\tabcolsep]}l}{\textbf{\modelname}} & \textbf{0.259} & \textbf{0.074} & \textbf{0.214} & \textbf{0.182} & 0.131 & \textbf{0.233} & 0.238 & \multicolumn{1}{>{\columncolor{oursrow}[\tabcolsep][0pt]}c@{}}{\textbf{0.201}} \\
        \bottomrule
    \end{tabularx}
\end{table*}

%% file: tables/outer_loop_ablation.tex
\begin{table*}[!t]
\centering
\small
\caption{\textbf{Comparison with RL-based meta-training on Jericho (W-AUC).} All rows share the same Gemini-3-Flash actor and Qwen3-4B meta-agent, trained with 350 rollouts (\modelname) vs.\ 1{,}920 (GRPO).}
\label{tab:outer_loop_ablation}
\begin{tabularx}{\textwidth}{@{}l*{8}{Y}@{}}
\toprule
\multirow{2}{*}{\textbf{Meta-Agent Training}}
& \multicolumn{4}{c}{\textbf{ID}}
& \multicolumn{4}{c}{\textbf{OOD}} \\
\cmidrule(lr){2-5} \cmidrule(lr){6-9}
& \textbf{Detective}
& \textbf{Zork~1}
& \textbf{Temple}
& \textbf{Avg.}
& \textbf{Balances}
& \textbf{Library}
& \textbf{Zork~3}
& \textbf{Avg.} \\
\midrule
Base (untrained)
& 0.198 & 0.110 & 0.129 & 0.146
& 0.105 & 0.217 & 0.181 & 0.168 \\

RL outer loop (GRPO)
& 0.219 & 0.117 & 0.114 & 0.150
& 0.121 & 0.172 & 0.229 & 0.174 \\

GEPA outer loop
& 0.218 & 0.085 & \textbf{0.154} & 0.152
& 0.124 & 0.252 & 0.219 & 0.198 \\

\rowcolor{oursrow}
\multicolumn{1}{@{}>{\columncolor{oursrow}[0pt][\tabcolsep]}l}{\modelname outer loop}
& \textbf{0.235} & \textbf{0.127} & 0.129 & \textbf{0.164}
& \textbf{0.150} & \textbf{0.300} & \textbf{0.247} & \multicolumn{1}{>{\columncolor{oursrow}[\tabcolsep][0pt]}c@{}}{\textbf{0.232}} \\
\bottomrule
\end{tabularx}
\end{table*}

%% file: tables/strategies.tex
\begin{figure}[t]
\centering
\begin{tcbraster}[raster columns=3, raster equal height, raster column skip=6pt,
  enhanced, boxrule=0.5pt, arc=2pt,
  colback=oursrow!45, colframe=stratframe, coltitle=white,
  fonttitle=\bfseries\small, fontupper=\small\itshape\raggedright,
  left=5pt, right=5pt, top=4pt, bottom=4pt, halign title=center]
\begin{tcolorbox}[title=Meta-level credit assignment]
``If prohibited behavior repeats, diagnose guidance-compliance failure rather than repeating the same advice.''
\end{tcolorbox}
\begin{tcolorbox}[title=Verified knowledge retention]
``Credit useful actor improvisations and promote them only under the source conditions where they succeeded.''
\end{tcolorbox}
\begin{tcolorbox}[title=Bounded exploration]
``With a strong baseline, place uncertain tests where failure sacrifices as little reliable score as possible.''
\end{tcolorbox}
\end{tcbraster}
\caption{\textbf{Excerpts from the optimized meta-prompts}, one per learned strategy (full prompts in Appendix~\ref{app:optimized_prompts}).}
\label{fig:strategies}
\end{figure}

%% file: tables/efficiency.tex
\begin{table}[t]
\centering
\small
\caption{\textbf{Inference efficiency comparison.} Latency and LLM usage for six-episode TTL sessions on Jericho. Numbers are averages over 3 runs.}
\label{tab:efficiency}
\begin{tabular}{lcccc}
\toprule
\textbf{Method} & \textbf{Wall-clock (s)} & \textbf{Total LLM calls}
& \textbf{Adaptation LLM calls} & \textbf{Tokens} \\
\midrule
Static       & $593.2 \pm 61.3$  & 231.0 & 0  & 196k \\
Reflexion    & $594.3 \pm 69.6$  & 218.0 & 0  & 270k \\
Memory Agent & $710.1 \pm 30.1$  & 262.0 & 12 & 311k \\
AWM & $840.1 \pm 27.2$  & 255.7 & 5 & 334k \\
TextGrad     & $1205.5 \pm 7.9$  & 264.0 & 30 & 384k \\
EvoPrompt    & $1266.5 \pm 29.4$ & 447.0 & 5  & 448k \\
EvoTest      & $759.1 \pm 68.1$  & 254.0 & 5  & 181k \\
\midrule
\modelname   & $860.9 \pm 66.2$  & 204.0 & 5  & 364k \\
\bottomrule
\end{tabular}
\end{table}

%% file: tables/backbones_jericho.tex
\begin{table}[h]
\centering
\small
\caption{\textbf{Results on Jericho with a GLM-5.3-Flash meta-agent.} All methods use a frozen Gemini-3-Flash actor. Each value is averaged over the ID or OOD games. Best results are in \textbf{bold} and second-best results are \underline{underlined}.}
\label{tab:backbones_jericho}
\begin{tabular}{@{}lcccc@{}}
\toprule
& \multicolumn{2}{c}{ID} & \multicolumn{2}{c}{OOD} \\
\cmidrule(lr){2-3} \cmidrule(l){4-5}
\textbf{Method} & Avg. Score & W-AUC & Avg. Score & W-AUC \\
\midrule
Naive Meta-Agent & 0.213 & 0.215 & 0.274 & 0.274 \\
AWM & 0.232 & 0.246 & 0.325 & 0.320 \\
EvoTest & \underline{0.285} & \underline{0.295} & \underline{0.334} & \underline{0.323} \\
\midrule
\rowcolor{oursrow}
\multicolumn{1}{@{}>{\columncolor{oursrow}[0pt][\tabcolsep]}l}{\modelname}
& \textbf{0.327} & \textbf{0.379} & \textbf{0.343}
& \multicolumn{1}{>{\columncolor{oursrow}[\tabcolsep][0pt]}c@{}}{\textbf{0.371}} \\
\bottomrule
\end{tabular}
\end{table}

%% file: tables/backbones_webarena.tex
\begin{table}[h]
\centering
\small
\caption{\textbf{Results on WebArena-Lite with a GLM-5.3-Flash meta-agent.} All methods use a frozen Gemini-3-Flash actor. Each value is averaged over the ID or OOD domains. Best results are in \textbf{bold} and second-best results are \underline{underlined}.}
\label{tab:backbones_webarena}
\begin{tabular}{@{}lcccc@{}}
\toprule
& \multicolumn{2}{c}{ID} & \multicolumn{2}{c}{OOD} \\
\cmidrule(lr){2-3} \cmidrule(l){4-5}
\textbf{Method} & Avg. Score & W-AUC & Avg. Score & W-AUC \\
\midrule
Naive Meta-Agent & 0.589 & 0.622 & \underline{0.746} & \underline{0.767} \\
AWM & \underline{0.611} & \underline{0.644} & 0.722 & 0.731 \\
EvoTest & 0.600 & 0.628 & 0.728 & 0.735 \\
\midrule
\rowcolor{oursrow}
\multicolumn{1}{@{}>{\columncolor{oursrow}[0pt][\tabcolsep]}l}{\modelname}
& \textbf{0.700} & \textbf{0.750} & \textbf{0.806}
& \multicolumn{1}{>{\columncolor{oursrow}[\tabcolsep][0pt]}c@{}}{\textbf{0.820}} \\
\bottomrule
\end{tabular}
\end{table}